\documentclass[journal]{IEEEtran}
\usepackage{tabularx}
\usepackage{booktabs}
\usepackage{dg_usepackage}
\usepackage{dg_def}

\ifCLASSINFOpdf
            \else
                  \fi

\begin{document}
\title{Learning Deep Gradient Descent Optimization for Image Deconvolution}

\author{Dong Gong, Zhen Zhang, Qinfeng Shi, Anton van den Hengel, Chunhua Shen, and Yanning Zhang
\thanks{This work was supported in part by ARC Discovery Project under Grant DP160100703 and DP200103797. Y. Zhang was supported by the Chang Jiang Scholars Program of China under Grant 100017GH030150 and 15GH0301.}
\thanks{D. Gong, Z. Zhang, Q. Shi, A. van den Hengel and C. Shen are with The University of Adelaide, SA 5005, Australia (e-mail: edgong01@gmail.com; zhen@zzhang.org; javen.shi@adelaide.edu.au; anton.vandenhengel@adelaide.edu.au; chunhua.shen@adelaide.edu.au). }
\thanks{Y. Zhang is with the School of Computer Science and Engineering, Northwestern Polytechnical University, Xi'an, China, 710129
(e-mail: ynzhang@nwpu.edu.cn).}
\thanks{D. Gong and Z. Zhang contributed equally to this work. Partial work was done when D. Gong was with Northwestern Polytechnical University. }
}

\markboth{IEEE Transactions on Neural Networks and Learning Systems}{Shell \MakeLowercase{\textit{et al.}}: Bare Demo of IEEEtran.cls for Journals}

\maketitle
\begin{abstract}
As an integral component of blind image deblurring, non-blind deconvolution removes image blur with a given blur kernel, which is essential but difficult due to the ill-posed nature of the inverse problem. The predominant approach is based on optimization subject to regularization functions that are either manually designed, or learned from examples. Existing learning based methods have shown superior restoration quality but are not practical enough due to their restricted and static model design. They solely focus on learning a prior and require to know the noise level for deconvolution. We address the gap between the optimization-based and learning-based approaches by learning a universal gradient descent optimizer. We propose a Recurrent Gradient Descent Network (RGDN) by systematically incorporating deep neural networks into a fully parameterized gradient descent scheme. A hyper-parameter-free update unit shared across steps is used to generate updates from the current estimates, based on a convolutional neural network. By training on diverse examples, the Recurrent Gradient Descent Network learns an implicit image prior and a universal update rule through recursive supervision. The learned optimizer can be repeatedly used to improve the quality of diverse degenerated observations. The proposed method possesses strong interpretability and high generalization. Extensive experiments on synthetic benchmarks and challenging real-world images demonstrate that the proposed deep optimization method is effective and robust to produce favorable results as well as practical for real-world image deblurring applications.

\end{abstract}


\IEEEpeerreviewmaketitle
\section{Introduction}

\IEEEPARstart{I}{mage} deconvolution, also known as image deblurring, aims to recover a sharp image from an observed blurry image.
The blurry image $\by\in \mbR^m$ is usually modeled as a convolution of a latent image $\bx\in\mbR^n$ and a blur kernel $\bk\in \mbR^l$:
\begin{equation}
\by = \bk * \bx + \bn,
\label{eq:blur_model}
\end{equation}
where $*$ denotes the convolution operator, $\bn\in \mbR^m$ denotes an i.i.d. white Gaussian noise term with unknown standard deviation (\ie noise level).
Given a blurry image $\by$ and the corresponding blur kernel $\bk$, the task of recovering the sharp image $\bx$ is referred to as \emph{(non-blind) image deconvolution}, which is often used as a subcomponent of blind image deblurring \cite{pan2014text,xu2010two,gong2016motion}.

\par
Single image deconvolution is challenging and mathematically ill-posed
due to the unknown noise and the loss of the high-frequency information.
Many conventional methods resort to different natural image priors based on manually designed empirical statistics (\eg sparse gradient prior \cite{levin2007image,krishnan2009fast,wang2008new}) or learned generative models (\eg Gaussian mixture models (GMMs) \cite{zoran2011epll}), which usually lead to non-convex and time-consuming optimization.
The optimization algorithms are used to iteratively update the images based on the priors and the imaging model in \eqref{eq:blur_model}.
For efficiency, discriminative learning methods \cite{schmidt2014csf,schuler2013mlp,xu2014deep} are investigated to learn mapping functions from blurred observation to the sharp image, which are usually restricted to specific blur kernels and noise levels, however.

\par
Due to the successes in many computer vision applications, deep neural networks (DNNs) have been used more frequently for learning image restoration models \cite{xu2014deep,zhang2016FCN,chang2017one,zhang2017learning,liu2018learning}.
Since it is impractical to directly apply end-to-end DNNs to the deconvolution for different blur kernels \cite{xu2014deep}, many approaches resort to unrolling an optimization algorithm as a static cascade scheme with a fixed number of steps in which specific neural networks are integrated into different steps \cite{zhang2016FCN,zhang2017learning,schuler2013mlp,kruse2017FFT}. The DNN components are usually model the operators only corresponding to the priors/regularizers (\eg proximal projectors \cite{zhang2017learning,chang2017one}).
In these static model structures, the DNN based operators in each step are learned specifically for the intermediate output from the previous step. As a result, these models usually require customized training for specific noise levels \cite{zhang2016FCN,schuler2013mlp} or manually parameter tuning (that reflects the unknown noise level) for a specific blurred image (in testing) \cite{kruse2017FFT,chang2017one,zhang2017learning}, limiting their applications in practice.
Although the learning based methods have applied the optimization schemes as an interface to the deconvolution application, they are restricted to learn a static mapping function and overlook the dynamic characteristics in the optimization process.

\par
We address the above issues by learning a universal optimizer for image deconvolution. Specifically, we propose \emph{Recurrent Gradient Descent Network (RGDN)}, a recurrent DNN architecture derived from gradient descent optimization methods.
The RGDN iteratively updates the unknown variable $\bx$ using a universal image updating unit, which mimics the gradient descent optimization process.
To achieve this, we parametrize and learn a universal gradient descent \emph{optimizer}, which can be repeatedly used
to update $\bx$ based on its previous updates.
Unlike previous methods \cite{zhang2017learning,zhang2016FCN,chang2017one,kruse2017FFT} only focusing on image prior learning, we parametrize and learn all main operations of a general gradient descent algorithm, including the gradient of a free-form image prior, based on CNNs (see Fig. \ref{fig:rgdn_overview}). In previous methods \cite{zhang2017learning,zhang2016FCN}, the CNNs are mainly used as a denoiser on the image gradients in some splitting technique based optimization methods. 
{In the implementation of the proposed learnable optimizer, we observe that incorporating the standard optimization algorithm into the deep neural network design is beneficial since it can utilize the structure of the problem more effectively. 
The proposed model learns not only the optimization processes but also the items associating to the regularizer, reflecting an image prior. }
Moreover, the optimizer shared across steps is trained to dynamically handle different updating statues, which is more flexible and general to handle the observation with different blur and noise levels. Given input images with different levels of degenerations, the learned optimizer can adaptively obtain high-quality results via different numbers of iterations (see Fig. \ref{fig:inter} and \ref{fig:psnr_vs_ite}).

\par
To summarize, the main contributions of this paper are:
\begin{itemize}[topsep=-1pt] \item We learn an optimizer for image deconvolution by fully parameterizing the general gradient descent optimizer, instead of learning only image priors \cite{zoran2011epll,schmidt2010generative} or the prior-related operators \cite{zhang2017learning,kruse2017FFT,chang2017one}.
The integration of trainable DNNs and the fully parameterized optimization algorithm yields a parameter-free, effective and robust deconvolution method, making a substantial step towards the practical deconvolution for real-world images.

\item We propose a new discriminative learning model, \ie the RGDN, to  learn an optimizer for image deconvolution. The RGDN systematically incorporates a series of CNNs into the general gradient descent scheme. Benefiting from the parameter sharing and recursive supervision, RGDN tends to learn a universal and dynamic updating unit (\ie optimizer), which can be
iteratively applied arbitrary times to boost the performance on different observations,
making it a very flexible and practical method.

\item Training one RGDN model is able to handle various types of blur and noise.
Extensive experiments on both synthetic data and real images show that the parameter-free RGDN learned from a synthetic dataset can produce competitive or even better results against the other state-of-the-art methods requiring given/known noise level.
\end{itemize}

\section{Related Work}
Non-blind image deconvolution has been extensively studied in computer vision, signal processing and other related fields. We will only discuss the most relevant works. Existing non-blind deconvolution methods can be mainly categorized into two groups:
manually-designed conventional methods
and the learning based methods.

\noindent \textbf{Empirically designed non-blind deconvolution}
Many manually-designed approaches use empirical statistics on natural image gradients as the regularization or prior term \cite{wang2008new,krishnan2009fast,levin2007image}, such as the total variation (TV) regularizer \cite{wang2008new,mptvgong}, sparsity prior on second-order image gradients \cite{levin2007image} and approximate hyper-Laplacian distribution \cite{krishnan2009fast}.
Meanwhile, various optimization methods have been studied for solving image deconvolution problem, \eg alternating direction method of multipliers (ADMM) \cite{goldstein2014fast}.
These conventional methods are often sensitive to the parameter settings and may be computationally expensive.

\noindent \textbf{Learning-based non-blind deconvolution}
Rather than using manually-designed regularizers, some methods learn generative models from data as image priors \cite{zoran2011epll,schmidt2010generative,sun2014good}.
Zoran and Weiss \cite{zoran2011epll} propose a GMM-based image prior and a corresponding algorithm (EPLL) for deconvolution, which is further extended in \cite{sun2014good}. EPLL is effective but very computationally expensive. Schmidt \etal \cite{schmidt2010generative} train a Markov Random Field (MRF) based natural image prior for image restoration. Similar to the manually-designed priors, the learned priors also require well tuned parameters for specific noise levels.

\par
To improve efficiency, some approaches address deconvolution by directly learning a discriminative function \cite{schmidt2016cascades,schmidt2014csf,schuler2013mlp,kruse2017FFT,zhang2017learning}.
Schuler \etal \cite{schuler2013mlp} impose a regularized inversion of the blur in the Fourier domain and then remove the noise using a learned multi-layer perceptron (MLP). Schmidt and Roth \cite{schmidt2014csf} propose shrinkage fields (CSF), an efficent discriminative learning procedure based on a random field structure. Schmidt \etal \cite{schmidt2016cascades} propose an approach based on Gaussian conditional random field, in which the parameters are calculated through regression trees. 
{Chen \etal \cite{chen2015learning} proposed a diffusion network that integrates the learnable diffusion process into an iterative estimation scheme and can achieve high-quality results. However, this model merely focuses on modeling the image priors relying on the RBF based diffusion process and has to be specially trained for different noise levels.  }

\par
Deep neural networks have been studied as a more flexible and efficient approach for deconvolution. Xu \etal \cite{xu2014deep} train a CNN to restore the images with outliers in an end-to-end fashion, which requires a fine-tuning for every blur kernel.
As shown by the plug-and-play framework \cite{venkatakrishnan2013plug,heide2016proximal}, the variable splitting techniques \cite{geman1995nonlinear,goldstein2014fast} can be used to decouple the  restoration problem as a data fidelity term and a regularization term corresponding to a projector in optimization.
To handle the instance-specific blur kernel more easily, a series
of methods \cite{zhang2016FCN,zhang2017learning,chang2017one} learn a denoisor and integrate it into the optimization as the projector reflecting the regularization.
In \cite{zhang2016FCN}, a fully convolutional network (FCN) is trained to remove noise in image gradients to guide the image deconvolution, which has to be custom-trained for specific noise level.
Zhang \etal \cite{zhang2017learning} learn a set of CNN denoisors (for different noise levels) and plug them into a half-quadratic splitting (HQS) scheme for image restoration. Chang \etal \cite{chang2017one} learn a proximal operator with adversarial training as image prior.
Relying on HQS, Kruse \etal \cite{kruse2017FFT} learn a CNN-based prior term companying with an FFT-based deconvolution scheme.
These methods only focus on learning the prior/regularization term, and
the noise level is required to be known in the testing phase. 
In a recent work, Jin \etal \cite{jin2017noiseblind} propose a Bayesian framework for noise adaptive deconvolution.
{In recent work, Jin \etal \cite{jin2017noiseblind} propose a Bayesian framework for noise adaptive deconvolution by generalizing the model in \cite{chen2015learning}. Unlike the proposed method, it models the images with restricted features and activation functions and applies a fixed number iterations with specific parameters.}

\noindent \textbf{Other related works}
{Early works \cite{li2016learning,andrychowicz2016learning,ravi2016optimization,zheng2015conditional} have explored the general idea of \emph{learning to optimize} for different tasks, such as few-shot learning \cite{ravi2016optimization} and semantic segmentation \cite{zheng2015conditional}. Different optimization algorithms and iterative inference techniques are unrolled as deep learning models. For example, mean-field inference for conditional random fields is implemented as a recurrent neural network for image semantic segmentation \cite{zheng2015conditional}. In \cite{andrychowicz2016learning}, a coordinate-wise LSTM is trained to train neural networks for image classification by mimicking gradient descent methods. 
}
{Learning based deep models motivated by optimization algorithms have also been applied for image restoration tasks \cite{kobler2017variational,klatzer2016learning,gu2018integrating}. In \cite{klatzer2016learning}, a sequence of energy minimization modules is learned for joint demosaicing and denoising. Kobler \etal \cite{kobler2017variational} train variational networks for image reconstruction relying on the incremental proximal gradient methods. Gu \etal \cite{gu2018integrating} integrates local and non-local denoiser into the HQS framework as image priors for image restoration. In \cite{gu2017learning}, dynamically updated guidance is used to enhance depth images in a bi-level optimization framework.}
Deep neural networks with recurrent structures have also been studied in many other low-level image processing tasks, such as blind image deblurring \cite{wieschollek2017learning,kim2017online}, image super-resolution \cite{kim2016deeply}, and image filtering \cite{liu2016recfilter}.

\section{Recurrent Gradient Descent Network}
In this section, we will first briefly revisit the classical model-based non-blind deconvolution problem and the general gradient descent algorithm. We then propose the RGDN model with a fully parameterized gradient descent scheme. Finally, we discuss how to perform training and deconvolution with RGDN.

\par
We consider the common blur model in \eqref{eq:blur_model}, which can also be rewritten as
\begin{equation}
\by=\bA\bx+\bn,
\label{eq:blur_model_matrix}
\end{equation}
where $\bA\in \mbR^{n\times m}$ denotes the convolution matrix of $\bk$. 
{The matrix $\bA$ represents the convolution operation $\bx*\bk$ as the matrix-vector multiplication $\bA\bx$ with a similar definition in \cite{mptvgong}. Note that we slightly abuse the notation $\bk*\bx$ to denote the 2D convolution operation, although $\bx$ and $\bk$ are defined as the vector representations of the image and kernel, respectively. }

\subsection{Revisiting Gradient Descent for Non-blind Deconvolution}
Based on the blur model \eqref{eq:blur_model} and the common Gaussian noise assumption, given a blurry image $\by$ and the blur kernel $\bk$, the desired solution of the non-blind deconvolution should minimize a data fidelity term $f(\bx)=\frac{1}{2\lambda}\|\by-\bA\bx\|_2^2$, where the weighting term $\lambda>0$ reflects the noise level in $\by$. Considering the ill-posed nature of the problem, given a regularizer $\Omega(\bx)$, the non-blind deconvolution can be achieved by solving the minimization problem
\begin{equation}
  \min_{\bx}\frac{1}{2\lambda}\|\by-\bA\bx\|_2^2 + \gamma\Omega(\bx),
  \label{eq:devon_min_prob}
\end{equation}
where the regularizer $\Omega(\bx)$ corresponds to the image prior, and the weighting term $\gamma\geq 0$ controls the strength of the regularization.
Generally, $\Omega(\bx)$ can be in any form, such as the classical choice TV regularizer \cite{wang2008new} or an arbitrary learning-based free-form regularizer.

\par
Although the optimization algorithms with high level abstractions (\eg proximal algorithm \cite{parikh2014proximal}) are often used for problem \eqref{eq:devon_min_prob} \cite{krishnan2009fast,zoran2011epll,gong2017mpgl}, to show the potential of the proposed idea,
 we start from the gradient descent method sitting in a basic level.
 Let $t$ denote the step index.
The vanilla gradient descent solves for $\widehat{\bx}$ (\ie an estimate of $\bx$) via a sequence of updates:
\begin{equation}
\begin{split}
\bd^t &=- (\grad f(\bx^t) + \gamma \grad \Omega(\bx^t)),\\
\bx^{t+1} &= \bx^t + \alpha^t \bd^t, \\
\end{split}
\label{eq:vanilla_gd}
\end{equation}
where $\bd^t$ denotes the descent direction, $\alpha^t$ denotes the step length, and $\grad f(\bx^t)$ and $\grad \Omega(\bx^t)$ denote the gradients of $f(\cdot)$ and $\Omega(\cdot)$ at step $t$.
In classic gradient methods, the step length $\alpha^t$ is usually determined
by an exact or approximate line search procedure
\cite{wright1999numerical}.
Specifically, for the deconvolution problem \eqref{eq:devon_min_prob},  $\grad f(\bx^t) = \frac{1}{\lambda}(\bA^\T\bA\bx^t-\bA^\T\by)$. Note that $\grad \Omega(\cdot)$ may also be a subgradient for some regularizers.

\par
To accelerate the optimization, we can scale the descent direction $\bd^t$ via a scaling matrix $\bD^t$ using the curvature information, which
can be determined
by different ways. For example, $\bD^t$ is the inverse Hessian matrix (or an approximation) when the second order information of the objective is used \cite{wright1999numerical}. We thus arrive a general updating equation at step $t$:
\begin{equation}
  \bx^{t+1} = \bx^t - \alpha^t\bD^t \left({1}/{\lambda}(\bA^\T\bA\bx^t - \bA^\T\by) + \gamma\grad \Omega(\bx^t) \right).
  \label{eq:final_gd_update}
\end{equation}
Given an initialization $\bx^0$, a general gradient descent solves problem \eqref{eq:devon_min_prob} by repeating the updating in \eqref{eq:final_gd_update} until some certain stopping conditions are achieved. The compact formulation in \eqref{eq:final_gd_update} offers an advantage to learn a universal parametrized optimizer.

\begin{figure*}[!t]
\centering
\begin{minipage}[b]{1\textwidth}
\centerline{
\begin{overpic}[trim=1 1 1 1, clip, width=0.9\textwidth]
{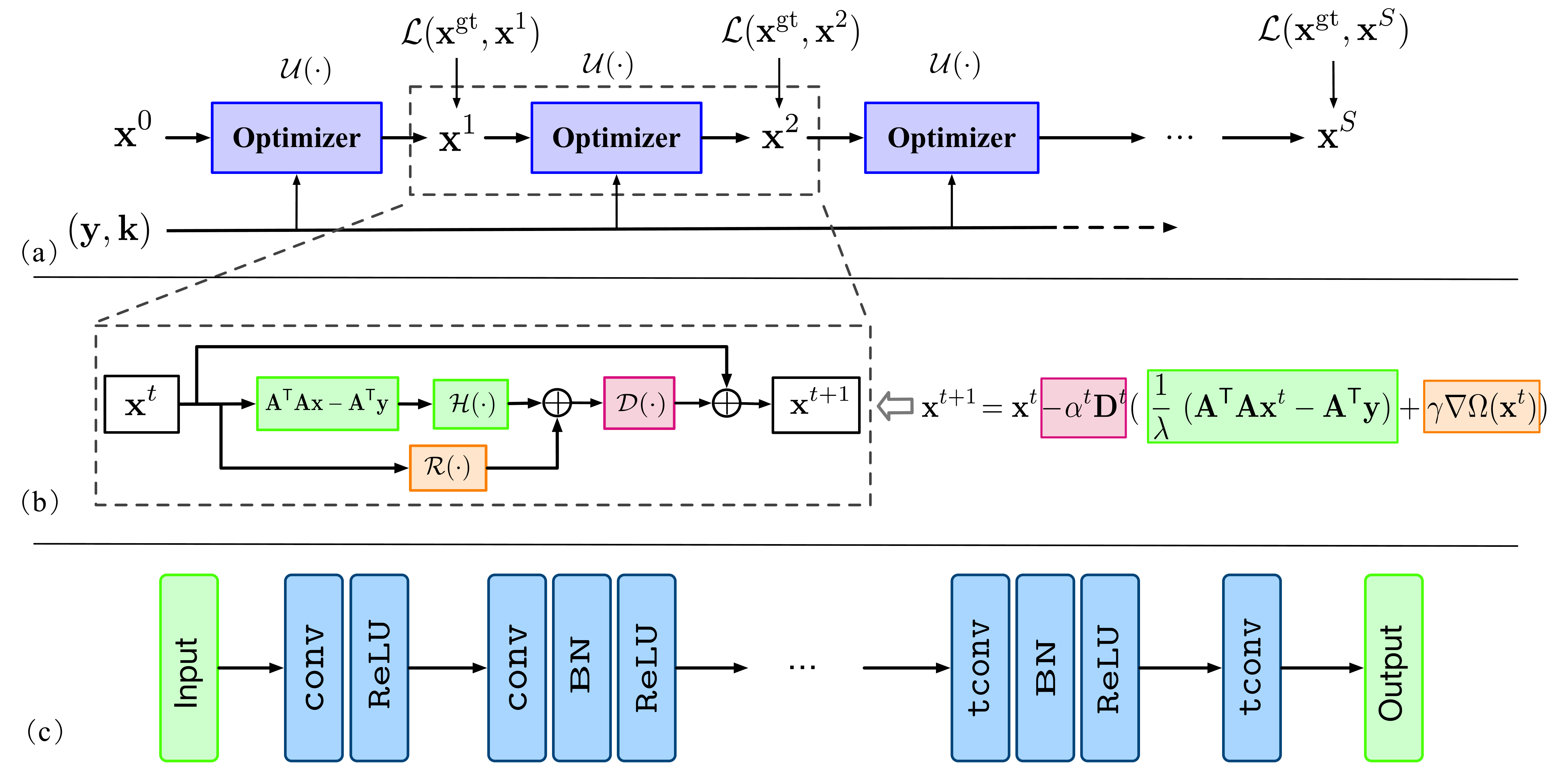}
\end{overpic}}
\end{minipage}
\caption{(a) The overall architecture of our RGDN. Given a blurry image $\by$ and the corresponding blur kernel $\bk$, the optimizer (\ie Gradient Descent Unit, GDU) $\cU(\cdot)$ produces a new estimate $\bx^{t+1}$ from the estimate from previous step $\bx^{t}$.
Note a universal optimizer is used for all steps with \textbf{shared parameters}. (b) The structure of the optimizer $\cU(\cdot)$. Each colored block of the optimizer (Left) corresponds to an operation in the classical gradient descent method (Right). (c) $\cR(\cdot)$, $\cH(\cdot)$ and $\cD(\cdot)$ share a common architecture of a CNN block with different parameters to be learned. Both the input and output (for the optimizer and all subnetworks) are $H\times W \times C$ tensors, where $C$ is the number of channels of the input image $\by$.
}
\label{fig:rgdn_overview}
\end{figure*}

\subsection{Parameterization of the Gradient Descent}
Our final goal is to learn a mapping function $\cF(\cdot)$ that takes a blurry image $\by$ and the blur kernel $\bk$ as input and recovers the target clear image $\bx$ as $\widehat{\bx}=\cF(\by, \bk)$. We achieve this by learning a fully parameterized optimizer.

\par
Given $\bx^t$ from the previous step, the gradient descent in \eqref{eq:final_gd_update} calculates $\bx^{t+1}$ relying on several main operations including gradient (or derivative) calculation for data fidelity term $f(\cdot)$ and regularizer $\Omega(\cdot)$, calculation of scaling matrix $\bD^t$ and step length determination.
For enabling the flexibility for learning, we fully parameterize the gradient descent optimizer in \eqref{eq:final_gd_update}. To achieve this, we replace the main computation entities with a series of parameterized mapping functions.
Firstly, we let $\cR(\cdot)$ replace $\grad \Omega(\cdot)$ to supplant the gradient of the regularizer. It implicitly plays as an image prior.
Considering that the noise level in $\by$ is unknown and hard to estimate in a prior, a predefined $\lambda$ is insufficient in practice.
We then define an operator $\cH(\cdot)$ to handle the unknown noise and the varying estimation error (in $\bx^t$) by adjusting $\bA^\T\bA\bx^t-\bA^\T\by$. $\cH(\cdot)$ implicitly tunes $\lambda$ adaptively.
Finally, we define $\cD(\cdot)$ as a functional operator to replace $\bD^t$ in each step to control the descent direction (\ie $\bd^t$).
$\cR(\cdot)$ and $\cD(\cdot)$ absorb the trade-off weight $\gamma$ and the step length $\alpha^t$, respectively. As shown in Fig. \ref{fig:rgdn_overview} (b), by replacing the calculation entities in \eqref{eq:final_gd_update} with the mapping functions introduced above, the gradient descent optimizer at each step can be formulated as:
\begin{equation}
\begin{split}
\bx^{t+1} & = \cU(\bx^t, \bk, \by) = \bx^t + \cG(\bx^t, \bk, \by)\\
&= \bx^t + \cD\left(\cR(\bx^t) + \cH(\bA^\T\bA\bx^t - \bA^\T\by)\right),
\end{split}
\label{eq:gd_para_update}
\end{equation}
where $\cU(\cdot)$ denotes the parametrized gradient descent optimizer, and $\cG(\cdot)$ denotes the gradient generator consisting of $\cR(\cdot)$, $\cH(\cdot)$ and $\cD(\cdot)$.
Given an initial $\bx^0$ (\eg letting $\bx^0=\by$), we can formulate the whole estimation model $\cF(\cdot)$ as
\begin{equation}
\begin{split}
\cF(\by,\bk; \Theta) & = \cU\circ  \cdots \circ \cU(\bx^0, \bk, \by)  = \cU^S(\bx^0, \bk, \by; \Theta),
\end{split}
\label{eq:gd_comp_fun}
\end{equation}
where $\circ$ denotes the composition operator, $\cU^S$ denotes a
the $S$-fold composition of $\cU(\cdot)$,
and $\Theta$ denotes the set of all parameters of $\cU(\cdot)$ (\ie the parameters of $\cR(\cdot)$, $\cH(\cdot)$ and $\cD(\cdot)$).
$\cU^S$ means the optimizer $\cU$ is performed $S$ times.
{In each iteration, the optimizer calculates once the gradient of data fitting term $\bA^\T\bA\bx-\bA^\T\by$. 
We use the matrix-vector formulation to simplify the representation and implement it using convolution operations for efficiency. Specifically, given any image $\bx$ and a blur kernel $\bk$ (or the equivalent matrix operator $\bA$), we can implement $\bA\bx$ and $\bA^\T\bx$ as $\bk*\bx$ and $\bar{\bk}*\bx$, respectively, where $\bar{\bk}$ denotes the blur kernel obtained by rotating $\bk$ counterclockwise by 180 degrees in the 2D plane. Thus $\bA^\T\bA\bx-\bA^\T\by$ can be implemented as $\bar{\bk}*\bk*\bx-\bar{\bk}*\by$. }

\begin{figure}[htp]
\centering
\subfigure[Blurry image $\by$]{
\begin{minipage}[b]{.14\textwidth}
\centerline{
\begin{overpic}[trim=3 3 3 3, clip, width=1\textwidth]
{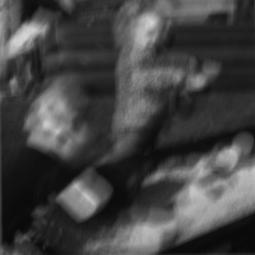}
\end{overpic}}
\end{minipage}
}
\subfigure[Input of $\cH(\cdot)$ ]{ \begin{minipage}[b]{.14\textwidth}
\centerline{
\begin{overpic}[trim=113 35 100 30, clip, width=1\textwidth]
{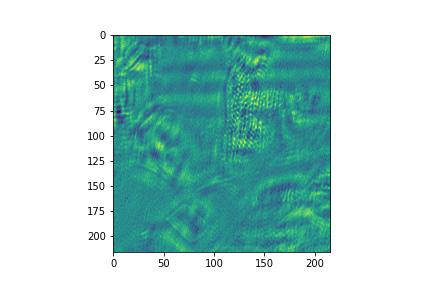}
\end{overpic}}
\end{minipage}
}
\subfigure[Output of $\cD(\cdot)$]{
\begin{minipage}[b]{.14\textwidth}
\centerline{
\begin{overpic}[trim=113 35 100 30, clip, width=1\textwidth]
{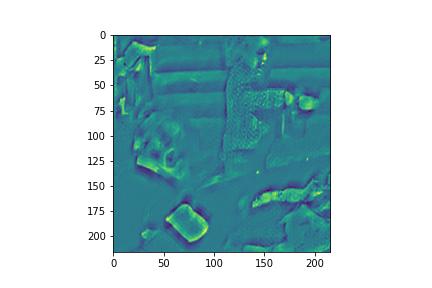}
\end{overpic}}
\end{minipage}
}
\subfigure[Output of $\cH(\cdot)$ ]{
\begin{minipage}[b]{.14\textwidth}
\centerline{
\begin{overpic}[trim=113 35 100 30, clip, width=1\textwidth]
{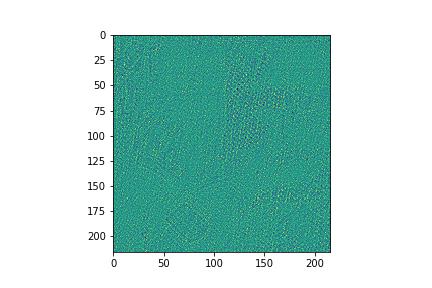}
\end{overpic}}
\end{minipage}
}
\subfigure[Output of $\!\cR(\cdot)$ ]{
\begin{minipage}[b]{.14\textwidth}
\centerline{
\begin{overpic}[trim=113 35 100 30, clip, width=1\textwidth]
{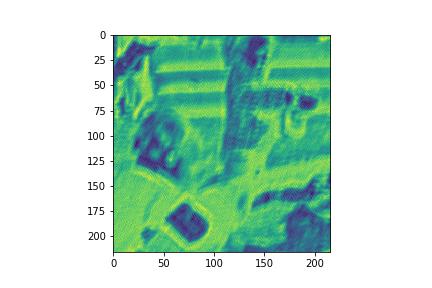}
\end{overpic}}
\end{minipage}
}
\subfigure[$\!\cR(\cdot)\!+\cH(\cdot)$]{
\begin{minipage}[b]{.14\textwidth}
\centerline{
\begin{overpic}[trim=113 35 100 30, clip, width=1\textwidth]
{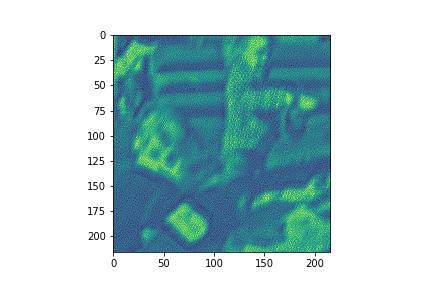}
\end{overpic}}
\end{minipage}
}
\caption{Visualization of each component of the learned gradient descent optimizer. The input of $\cH(\cdot)$ in (b) is the gradient from the loss function, \ie $\bA^\T\bA\bx-\bA^\T\by$. All images are scaled for visualization in pseudo color. The images are best viewed by zooming in.}
\label{fig:inter_RHD}
\end{figure}

\begin{figure}[htp]
{
\centering
\subfigure[Blurry image $\by$]{
\begin{minipage}[b]{.13\textwidth}
\centerline{
\begin{overpic}[trim =40 40 40 40, clip, width=1\textwidth]
{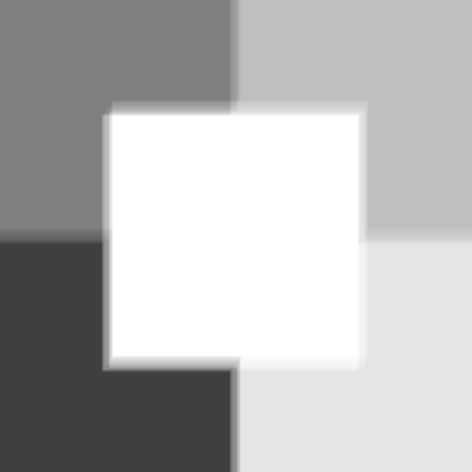}
\end{overpic}}
\end{minipage}
}
\hspace{-0.1cm}
\subfigure[Ground Truth $\bx^{\text{GT}}$]{
\begin{minipage}[b]{.13\textwidth}
\centerline{
\begin{overpic}[trim =60 60 60 60, clip, width=1\textwidth]
{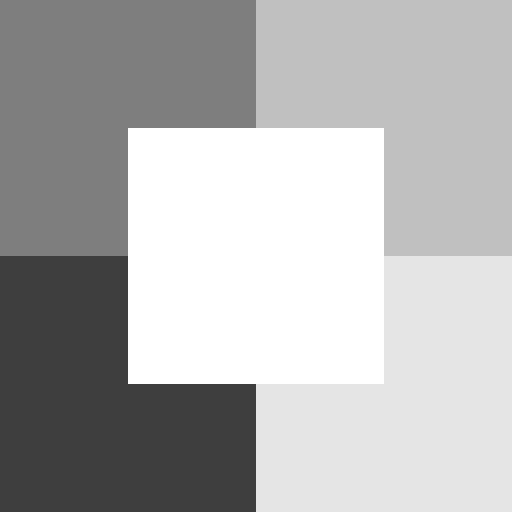}
\end{overpic}}
\end{minipage}
}
\hspace{-0.1cm}
\subfigure[Inter. estimate $\bx^3$]{
\begin{minipage}[b]{.13\textwidth}
\centerline{
\begin{overpic}[trim =40 40 40 40, clip, width=1\textwidth]
{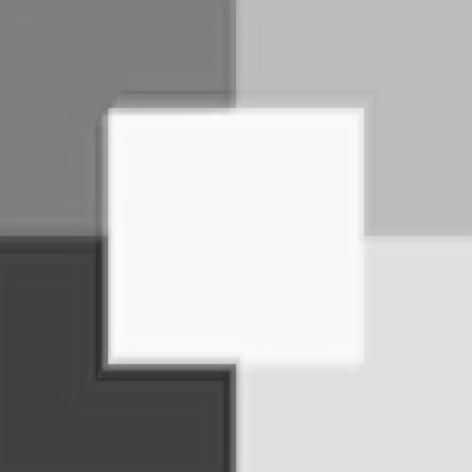}
\end{overpic}}
\end{minipage}
}
\hspace{-0.1cm}
\subfigure[$-\grad f(\bx^3)$]{
\begin{minipage}[b]{.13\textwidth}
\centerline{
\begin{overpic}[trim =10 10 10 10, clip, width=1\textwidth]
{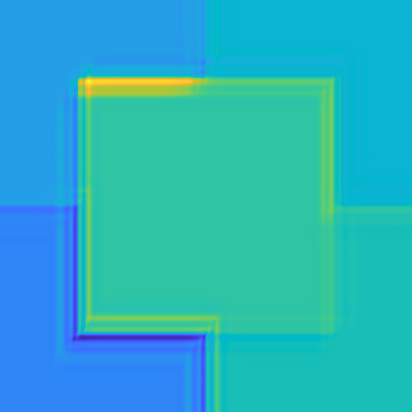}
\end{overpic}}
\end{minipage}
}
\hspace{-0.1cm}
\subfigure[$\cG(\bx^3,\bk,\by)$]{
\begin{minipage}[b]{.13\textwidth}
\centerline{
\begin{overpic}[trim =10 10 10 10, clip, width=1\textwidth]
{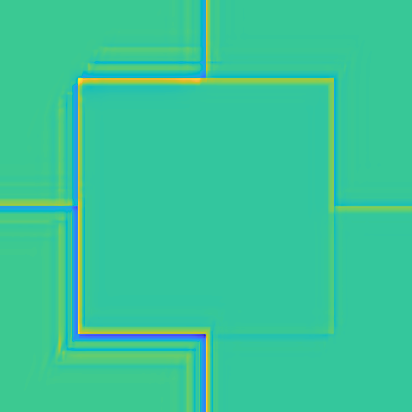}
\end{overpic}}
\end{minipage}
}
\hspace{-0.1cm}
\subfigure[$\mathbf{r}=\bx^{\text{GT}}-\bx^3$]{
\begin{minipage}[b]{.13\textwidth}
\centerline{
\begin{overpic}[trim =10 10 10 10, clip, width=1\textwidth]
{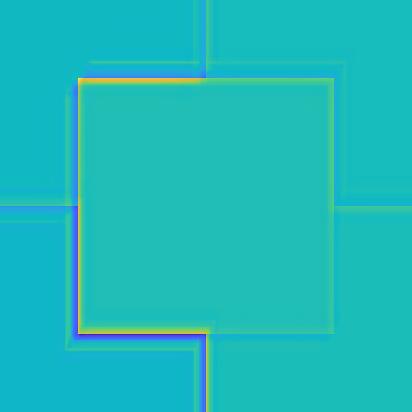}
\end{overpic}}
\end{minipage}
}
\caption{Visualization of the generated updating gradient on a toy image with elementary contents. (a) The input blurry image. (b) Ground truth image. (c) An intermediate estimate $\bx^t$ with $t=3$. (d) Gradient of the data fitting term, $-\grad f(\bx^3)=-(\bA^\T\bA\bx^3-\bA\by)$. (e) The generated gradient of the learned optimizer.
(f) The residual between the current estimate $\bx^3$ and the final target image, \ie the ground truth image $\bx^{\text{GT}}$, which can be seen as an ``ideal'' updating gradient to obtain the ground truth image with one step updating. The generated gradient in (e) is much more similar to the ``ideal'' gradient than the original gradient from data fitting term. 
Note that (d), (e), and (f) are visualized with scaling and pseudo-color.
The images are best viewed by zooming in.}
\label{fig:vis_grad}
}
\end{figure}

\subsection{The Structure of the RGDN}
\label{sec:RGDN_net_str}
We propose to formulate the model in equation~\eqref{eq:gd_comp_fun} as a Recurrent Gradient Descent Network (RGDN). Considering that the updates of $\bx^t$ from an iterative optimization scheme naturally compose a sequence of arbitrary length, we use a universal \emph{gradient descent unit (GDU)} to implement $\cU(\cdot)$ and apply it in all steps in a \emph{recurrent} manner (see Fig. \ref{fig:rgdn_overview} (a)).

\par
In the GDU, the gradient generator $\cG(\cdot)$ takes a current prediction $\bx^t$ of size $H \times W \times C$ and generates a gradient with the same size. In $\cU(\cdot)$, the subcomponents $\cR(\cdot)$, $\cH(\cdot)$ and $\cD(\cdot)$ play as mapping functions with a same size for input and output as well. 
Considering that CNNs with an encoder-decoder architecture have been commonly used to model similar mapping functions, we implement $\cR(\cdot)$, $\cH(\cdot)$ and $\cD(\cdot)$ using three CNNs with the same structure shown in Fig. \ref{fig:rgdn_overview} (c). 
Since finding the best structures for each subnetwork is not the main focus, we use the same structure as a default plain choice. Nevertheless, the three CNNs are trained with different parameters, resulting in different functions.
 We then construct the GDU by assembling the three CNNs according to the model in \eqref{eq:gd_para_update} (see Fig. \ref{fig:rgdn_overview} (b)).

\par
As shown in Fig. \ref{fig:rgdn_overview}, each trainable CNN consists
of 3 convolution layers ($\mathtt{conv}$) and 3 transposed convolution
($\mathtt{tconv}$) layers. Except for the first and the last layers,
each $\mathtt{conv}$ or $\mathtt{tconv}$ is followed by a batch
normalization ($\mathtt{BN}$) layer \cite{ioffe2015bn} and a ReLU activation function.
Following a widely used setting \cite{zhang2017learning}, the first $\mathtt{conv}$ is only followed by a ReLU activation function.
Apart from the last $\mathtt{tconv}$, we apply 64 $5\times 5$ convolution features for each $\mathtt{conv}$ and $\mathtt{tconv}$.
The last $\mathtt{tconv}$ maps the 64-channel intermediate features to a $C$-channel RGB output, where $C$ denotes the number of channels of the image. We set the stride size as 1 for all $\mathtt{conv}$ and $\mathtt{tconv}$.
Our contributions are agnostic to the specific implementation choice for the structure of each subnetwork corresponding to $\cR(\cdot)$, $\cH(\cdot)$ and $\cD(\cdot)$, respectively, which may be further tuned for better performance.

\par
\begin{figure*}[!t]
\centering
\hspace{-0.19cm}
\begin{minipage}[b]{.16\textwidth}
\centerline{
\begin{overpic}[trim=15 110 0 0,, clip, width=1\textwidth]
{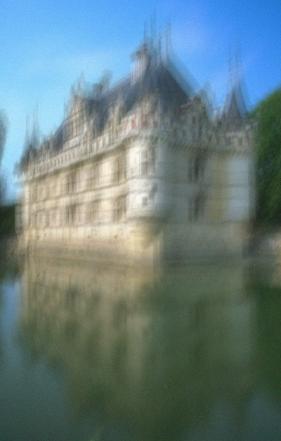}
  \put(68.5,187){ \includegraphics[width=0.1\textwidth]{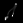}}
  \put(2,3){\footnotesize\color{white}{\bf (a) $\by$ and $\bk$}}
\end{overpic}}
\end{minipage}
\begin{minipage}[b]{.16\textwidth}
\centerline{
\begin{overpic}[trim=15 110 0 0, clip, width=1\textwidth]
{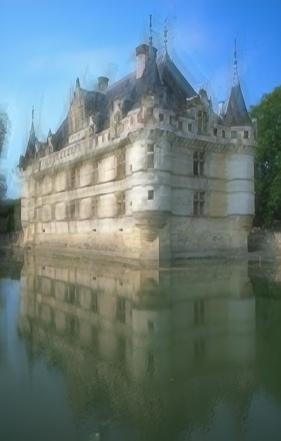}
  \put(2,3){\footnotesize\color{white}{\bf (b) Step \#3}}
\end{overpic}}
\end{minipage}
\begin{minipage}[b]{.16\textwidth}
\centerline{
\begin{overpic}[trim=15 110 0 0, clip, width=1\textwidth]
{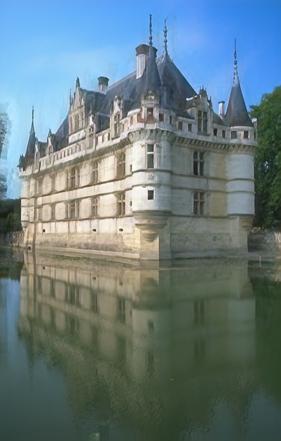}
  \put(2,3){\footnotesize\color{white}{\bf (c) Step \#20}}
\end{overpic}}
\end{minipage}
\begin{minipage}[b]{.16\textwidth}
\centerline{
\begin{overpic}[trim=15 110 0 0,clip, width=1\textwidth]
{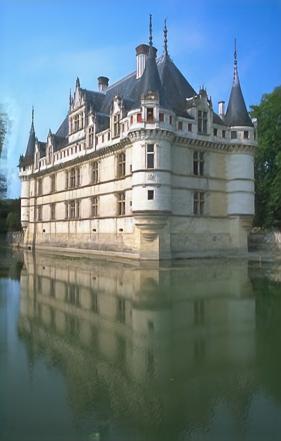}
  \put(2,3){\footnotesize\color{white}{\bf (d) Step \#30}}
\end{overpic}}
\end{minipage}
\begin{minipage}[b]{.16\textwidth}
\centerline{
\begin{overpic}[trim=15 110 0 0, clip, width=1\textwidth]
{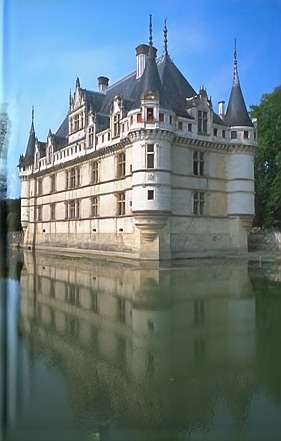}
  \put(2,3){\footnotesize\color{white}{\bf (e) Step \#40}}
\end{overpic}}
\end{minipage}
\begin{minipage}[b]{.16\textwidth}
\centerline{
\begin{overpic}[trim=35 130 20 20, clip, width=1\textwidth]
{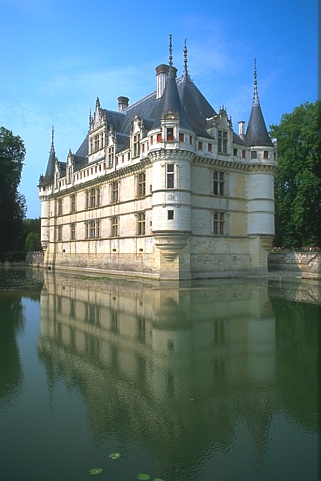}
  \put(2,3){\scriptsize\color{white}{\bf (f) Ground truth }}
\end{overpic}}
\end{minipage}\\
\vspace{0.1cm}
\begin{minipage}[b]{.16\textwidth}
\centerline{
\begin{overpic}[trim=0 5 10 5, clip, width=1\textwidth]
{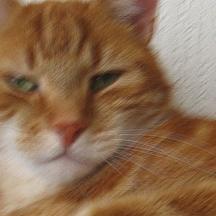}
  \put(70, 80){ \includegraphics[width=0.13\textwidth]{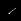}}
  \put(2,3){\footnotesize\color{white}{\bf (g) $\by$ and $\bx$}}
\end{overpic}}
\end{minipage}
\begin{minipage}[b]{.16\textwidth}
\centerline{
\begin{overpic}[trim=0 5 10 5, clip, width=1\textwidth]
{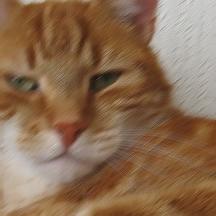}
  \put(2,3){\footnotesize\color{white}{\bf (h) Step \#1}}
\end{overpic}}
\end{minipage}
\begin{minipage}[b]{.16\textwidth}
\centerline{
\begin{overpic}[trim=0 5 10 5, clip, width=1\textwidth]
{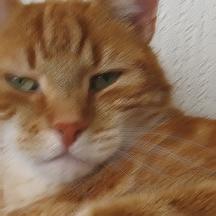}
  \put(2,3){\footnotesize\color{white}{\bf (i) Step \#2}}
\end{overpic}}
\end{minipage}
\begin{minipage}[b]{.16\textwidth}
\centerline{
\begin{overpic}[trim=0 5 10 5,clip, width=1\textwidth]
{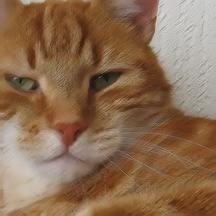}
  \put(2,3){\footnotesize\color{white}{\bf (j) Step \#3}}
\end{overpic}}
\end{minipage}
\begin{minipage}[b]{.16\textwidth}
\centerline{
\begin{overpic}[trim=0 5 10 5, clip, width=1\textwidth]
{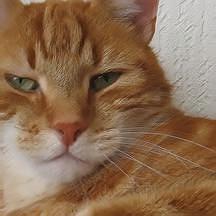}
  \put(2,3){\footnotesize\color{white}{\bf (k) Step \#5}}
\end{overpic}}
\end{minipage}
\begin{minipage}[b]{.16\textwidth}
\centerline{
\begin{overpic}[trim=0 5 10 5, clip, width=1\textwidth]
{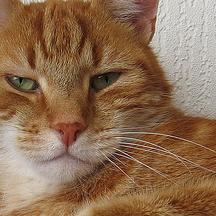}
  \put(2,3){\scriptsize\color{white}{\bf (l) Ground truth }}
\end{overpic}}
\end{minipage}
\vspace{-0.2cm}
\caption{Intermediate results of RGDN. (a) and (g) are the input blurry images $\by$ and the corresponding blur kernels $\bk$. (a) is an image with noise level 0.15\%. (g) is an image from the training data. Both the images and the kernels in (a) are not in the training set.
(b)-(e) are the intermediate results of the RGDN at the steps \#3, \#20, \#30 and \#40. (h)-(k) show the results on steps \#1-3 and \#5, since we perform 5 steps during training. (f) and (l) are the ground truth images.}
\label{fig:inter}
\end{figure*}

Towards learning a universal optimizer, the proposed RGDN shares parameters among the GDUs in all steps, which enables the optimizer (\ie the shared GDU) to see different states during the iterations. The learned optimizer can thus handle the dynamically varying states during the optimization. Combining with the recursive supervision introduced in the following, the learned universal optimizer can focus on \emph{improving} the quality of the current estimate in each update.
Training of the RGDN thus gives us flexibility to repeat the learned optimizer arbitrary times to approach the desired deconvolution results for different observations.
As a result, the proposed optimizer exhibits a strong generalization for handling images with different levels of degenerations, even beyond the training data (see Fig. \ref{fig:inter} and \ref{fig:psnr_vs_ite}).
In practice, we can stop the process relying on some stopping conditions as the classic iterative optimization algorithms.
Previous methods \cite{schmidt2014csf,kruse2017FFT} often truncate the classic iterative optimization algorithm with fixed step numbers and rigidly train different parameters to only process the images from previous steps. However, a static model with a fixed step number may not be suitable for all degenerated images. The previous models thus require the ground truth noise level as the input hyper-parameter and/or customized training for specific noise level, which limits the practicability.

\par
As shown in Fig. \ref{fig:rgdn_overview} (b), the three subnetworks corresponding to $\cR(\cdot)$, $\cH(\cdot)$ and $\cD(\cdot)$ are integrated together as the entities of a gradient descent and trained jointly. Although the network architecture is designed following the gradient descent process, for flexibility, we do not restrict each subnetwork to fit the exact intermediate output of the conventional gradient descent algorithm. The learned subnetworks thus may work beyond the functions of the conventional optimizer. 
{We observe that the learned optimizer can work stably and smoothly converges on various cases (as shown in Fig. \ref{fig:psnr_vs_ite}). Although the subcomponents $\cR(\cdot)$, $\cH(\cdot)$ and $\cD(\cdot)$ are not restricted to mimic the original optimization operations, they are implanted and trained in the optimization scheme, which enables the learned subnetworks to take benefits from the classic optimization scheme. Furthermore, the recursive supervisions on all steps push the universal optimizer to improve the image quality in each step.}
To give an intuitive understanding of the functions of the learned optimizer, we visualize the intermediate results produced by the subnetworks and show examples in Fig. \ref{fig:inter_RHD}. The input of $\cH(\cdot)$ in Fig. \ref{fig:inter_RHD} (b) is the gradient rising from the data fitting term, \ie $\bA^\T\bA\bx-\bA^\T\by$, which contains severe artifacts. 
The intermediate gradient generated by $\cH(\cdot)$ (in Fig. \ref{fig:inter_RHD} (d)) tends to remove the significant noise in the estimate. The subnetwork $\cR(\cdot)$ generate a descent direction only based on the current estimate. As shown in Fig. \ref{fig:inter_RHD} (c), the generated gradient explicitly handle the blurry boundary in the images and avoid to influence the details.
{Moreover, we synthesize a toy image with elementary contents and visualize the generated gradient descent direction more intuitively, as shown in Fig. \ref{fig:vis_grad}. 
It visualizes the generated gradient for updating $\bx$ in an intermediate step. Fig. \ref{fig:vis_grad} (f) shows the difference between the current estimate and the ground truth image $\bx^{\text{GT}}$ (shown in Fig. \ref{fig:vis_grad} (b)), which can be seen as an ideal updating gradient. 
The gradient of the data fitting term $f(\cdot)$ (shown in Fig. \ref{fig:vis_grad} (d)) is blurry and contains a lot of artifacts. 
The updating gradient generated by the proposed method is more similar to the ideal gradient (see Fig. \ref{fig:vis_grad} (e)). 
Fig. \ref{fig:vis_grad} shows that the gradient generated by the proposed optimizer is similar to the ``ideal'' gradient, and better than the ``original'' gradient from the data fitting term. }
{The visualizations in Fig. \ref{fig:inter_RHD} and \ref{fig:vis_grad} show that the gradient generated by the proposed learned optimizer is more effective than the simple gradient from the ordinary gradient $\bA^\T\bA\bx-\bA^\T\by$.
  }

\subsection{Learning an Optimizer via Training an RGDN}
\subsubsection{\textbf{Training loss}}
We expect to determine the best model parameter $\Theta$ that accurately estimates $\widehat{\bx}=\cF(\by,\bk; \Theta)$ through training on a given dataset $\{(\bx_i, \bk_i, \by_i)\}_{i=1}^N$.
We minimize the mean squared error (MSE) between the ground truth $\bx$ and the estimate $\widehat{\bx}$ over the training dataset:
\begin{equation}
\cL_{\text{MSE}}(\bx_i, \hat{\bx}_i; \Theta) = \|\bx_i-\widehat{\bx}_i\|_2^2.
\label{eq:loss_2}
\end{equation}
Inspired by \cite{zhang2016FCN,fan2017generic}, we also consider to minimize the gradient discrepancy in training:
\begin{equation}
\cL_{\text{grad}}(\bx_i, \hat{\bx}_i; \Theta) = \|\grad_v\bx_i-\grad_v\widehat{\bx}_i\|_1 + \|\grad_h\bx_i-\grad_h\widehat{\bx}_i\|_1,
\label{eq:loss_grad}
\end{equation}
where $\grad_v$ and $\grad_h$ denote the operators calculating the image gradients in the horizontal and vertical directions, respectively. The loss function in \eqref{eq:loss_grad} is expected to help to produce sharp images \cite{fan2017generic}.
For all experiments, the models are trained by minimizing the sum of $\cL_{\text{MSE}}(\cdot)$ and $\cL_{\text{grad}}(\cdot)$.

\subsubsection{\textbf{Recursive supervision and training objective}}
Instead of solely minimizing the difference between the ground truth and the output of the final step, we impose  recursive supervision \cite{kim2016deeply} that supervises not only the final estimate but also the outputs of the intermediate steps (\ie outputs of $\cU^t(\cdot)$'s, for $t\in [1, S)$). 
The recursive supervision directly forces the output of each step to approach the ground truth, which accelerates the training and enhances the performance (see Section \ref{sec:abl}). 
{If we apply supervision only on the last step, only the final optimization step provides information for training the optimizer, rendering inefficient gradient backpropagation through the recursive steps. 
The recursive supervision on the intermediate steps along the optimization trajectory allows us to train the optimizer on partial intermediate trajectories, which is similar to \cite{andrychowicz2016learning} and can help to tackle the issue. 
The supervisions on all steps help to train the optimizer that can achieve desired solution as fast as possible. }
Let $\widehat{\bx}_i^t=\cU^t(\bx_i^0, \by_i, \bk_i)$ denote the estimate of $\bx_i$ from the $t$-th step.
By averaging over all training samples and the steps, we have the whole training objective
{
\begin{equation}
\cL(\Theta) = \frac{1}{NS}\! \sum_{i=1}^N \sum_{t=1}^S \kappa_t \left( \cL_{\text{MSE}}(\bx_i, \widehat{\bx}_i^t; \Theta) + \tau\cL_{\text{grad}}(\bx_i, \widehat{\bx}_i^t; \Theta) \right),
\label{eq:obj_step_weight}
\end{equation}}
{where $\tau$ denotes the importance weight for the loss term on image gradients and the weights $\kappa_t, t=1,...,S$ denote the weights for the losses on different steps. 
In our implementation, we apply the default setting $\tau=1$ and $\kappa_t=1, t=1,..., S$ for simplicity, although there may exist a particular ``optimal'' setting for the weights. 
In Section \ref{sec:abl}, we conduct experiments to study the behaviors of different settings for $\tau$ and $\kappa_t$'s. 
}
As shown in Fig. \ref{fig:inter}, the learned optimizer steadily pushes the results close to the ground truth, which is consistent with the recursive supervision.

\subsubsection{\textbf{Implementation details}}
\label{sec:imple_detais}
Although the number of steps the RGBN takes is not bounded in principle, considering the training efficiency, we run the optimizer for 5 steps in training (\ie $S=5$). 
{As shown in experiments, benefiting from the parameter sharing and recursive supervision, the proposed learned optimizer can obtain sustained performance gain after running with the iterations more than that in training. This observation is consistent with the learnable optimizer based meta-learning method \cite{andrychowicz2016learning}. }

\par
For training, we randomly initialize the parameters of RGDN.
The training is carried out using a mini-batch Adam \cite{kinga2015adam} optimizer. We set the batch size and learning rate as 4 and $5\times 10^{-5}$, respectively.

\begin{figure*}[!t]
\small
  \centering
  \begin{center}
  \setlength{\tabcolsep}{1pt}
    \begin{tabularx}{0.95\textwidth}{cccccccccccc}
\multicolumn{2}{c}{\includegraphics[width=0.15\textwidth]{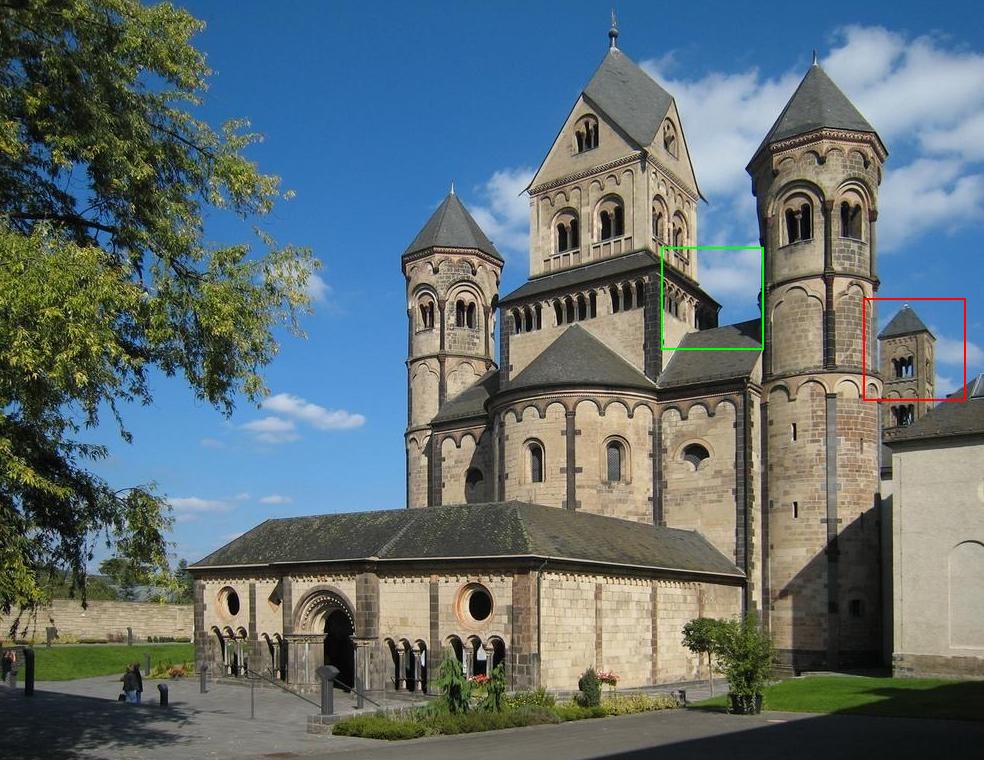}}
& \multicolumn{2}{c}{\includegraphics[width=0.15\textwidth]{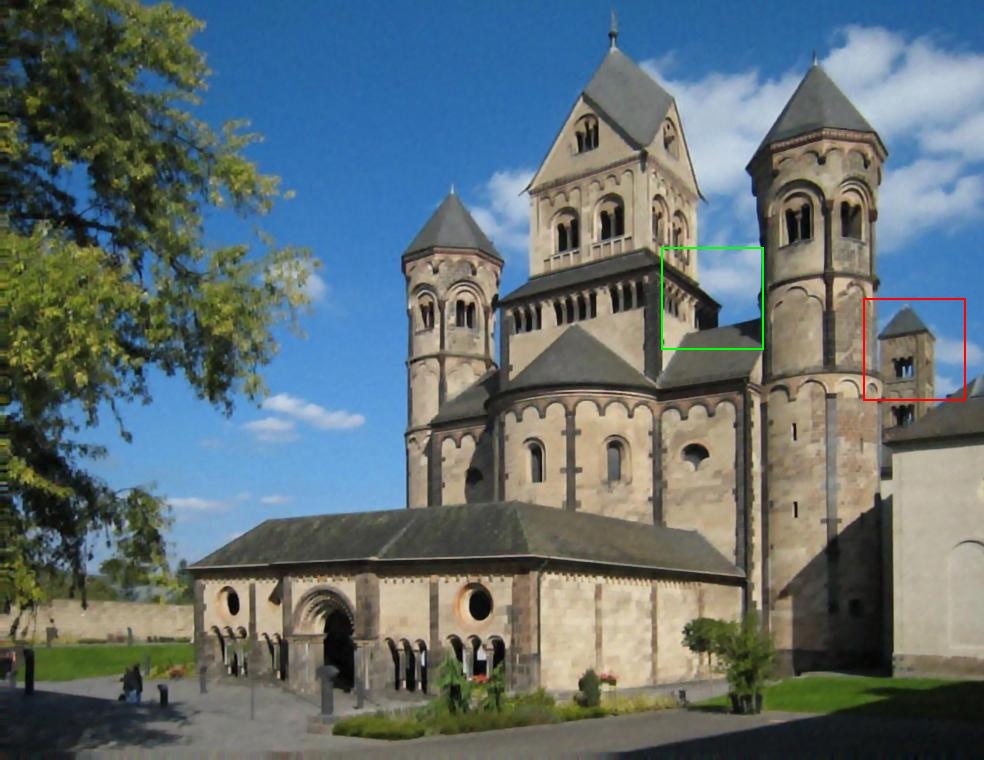}}
& \multicolumn{2}{c}{\includegraphics[width=0.15\textwidth]{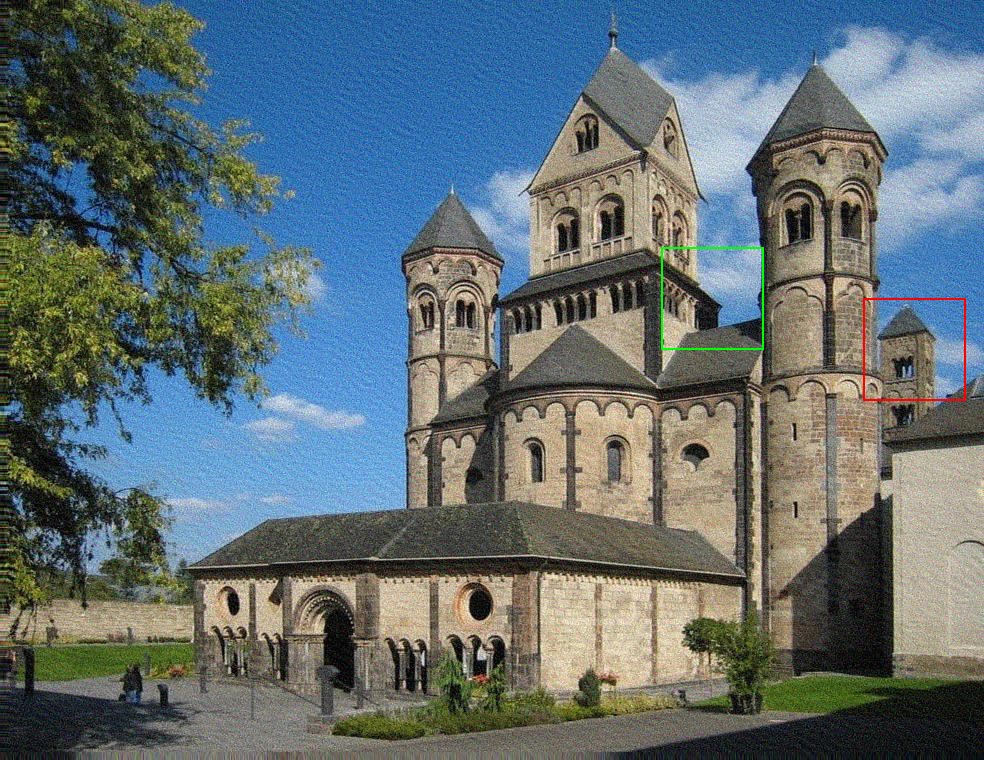}}
& \multicolumn{2}{c}{\includegraphics[width=0.15\textwidth]{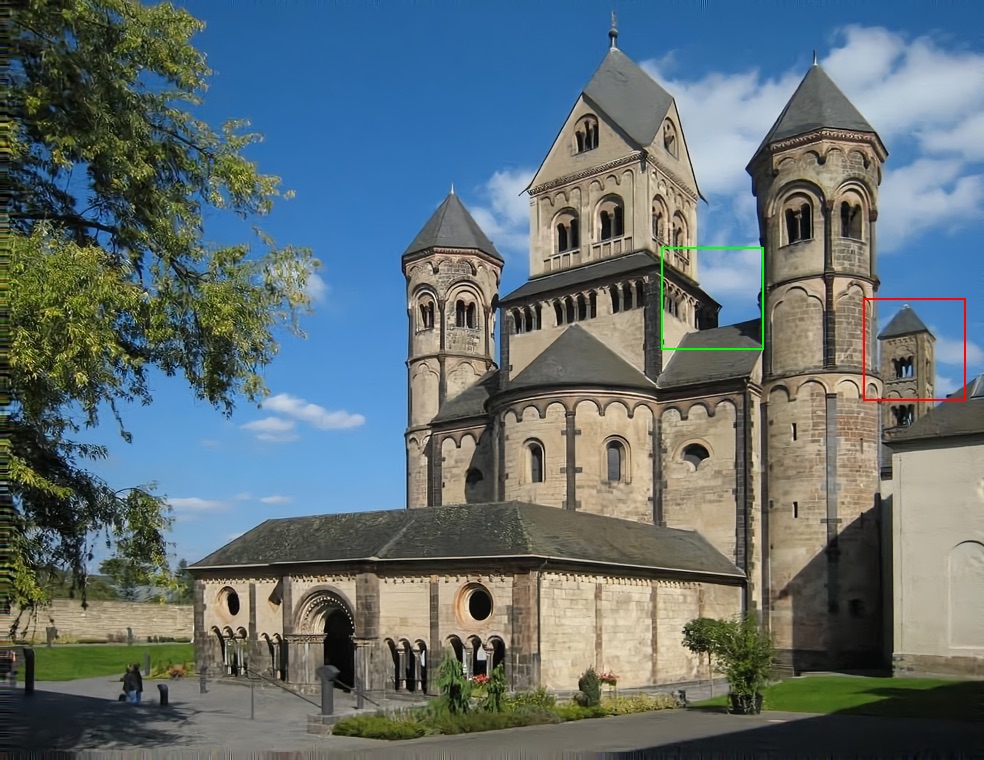}}
&     \multicolumn{2}{c}{\includegraphics[width=0.15\textwidth]{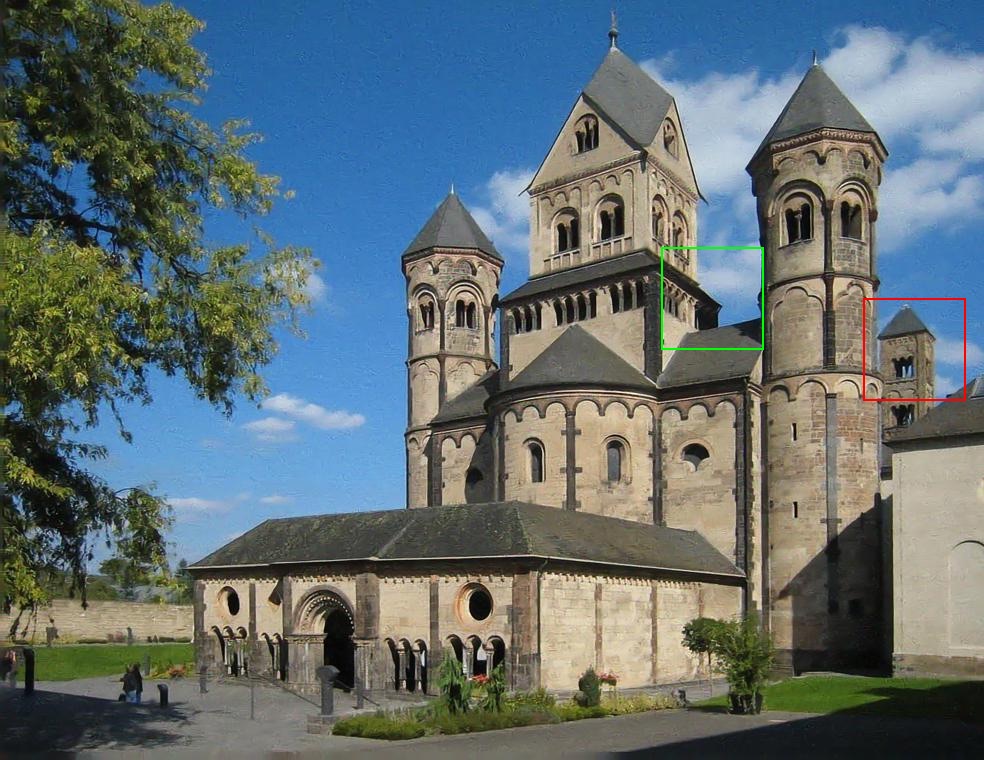}}                                                    &                \multicolumn{2}{c}{\includegraphics[width=0.15\textwidth]{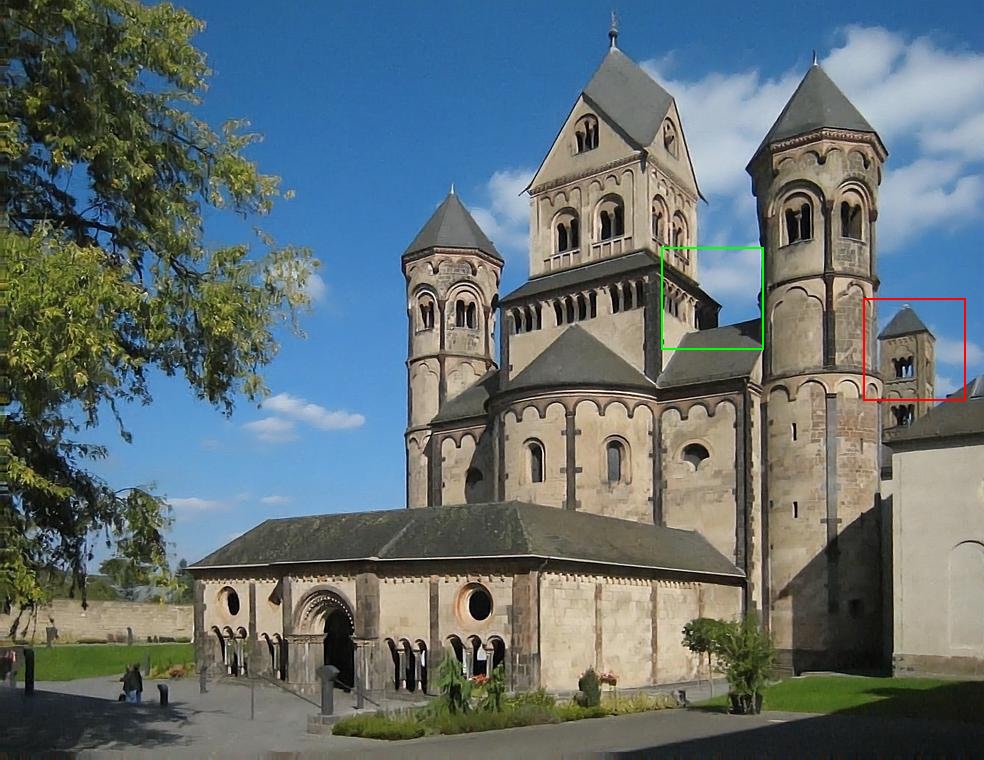}}\\
\includegraphics[width=0.075\textwidth]{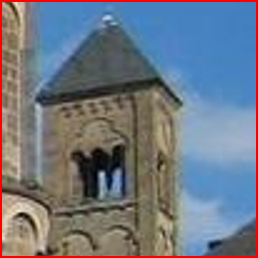}&
\includegraphics[width=0.075\textwidth]{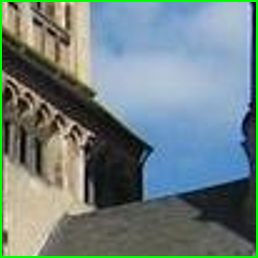}&
\includegraphics[width=0.075\textwidth]{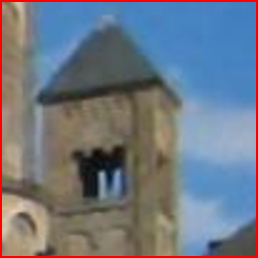}&
\includegraphics[width=0.075\textwidth]{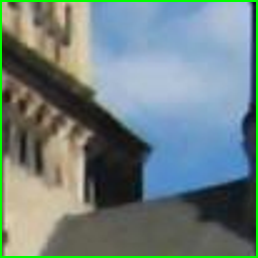}&
\includegraphics[width=0.075\textwidth]{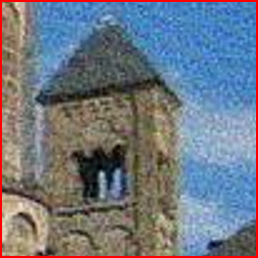}&
\includegraphics[width=0.075\textwidth]{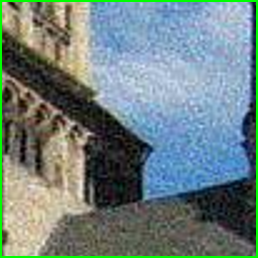}&
\includegraphics[width=0.075\textwidth]{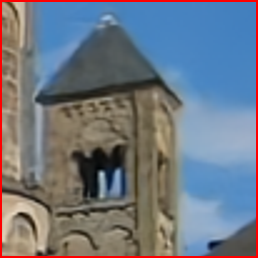}&
\includegraphics[width=0.075\textwidth]{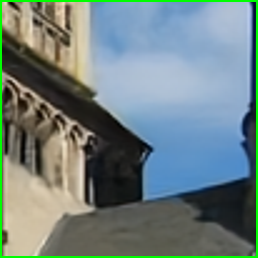}&

\includegraphics[width=0.075\textwidth]{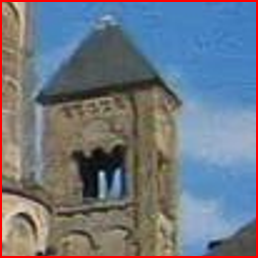}&
\includegraphics[width=0.075\textwidth]{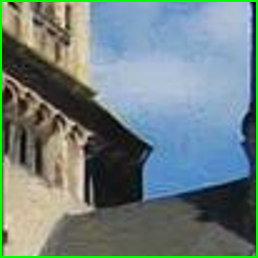}&
\includegraphics[width=0.075\textwidth]{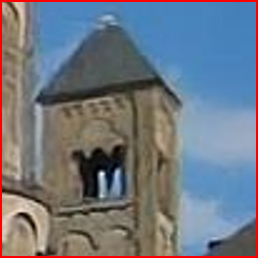}&
\includegraphics[width=0.075\textwidth]{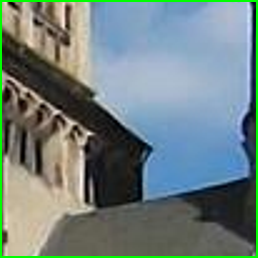}
\\
  \multicolumn{2}{c}{\includegraphics[width=0.15\textwidth]{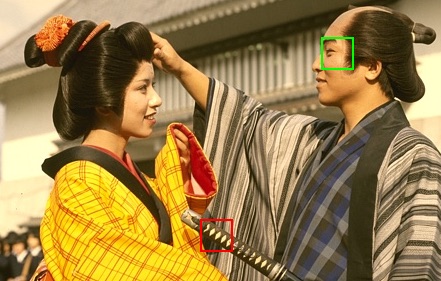}}
& \multicolumn{2}{c}{\includegraphics[width=0.15\textwidth]{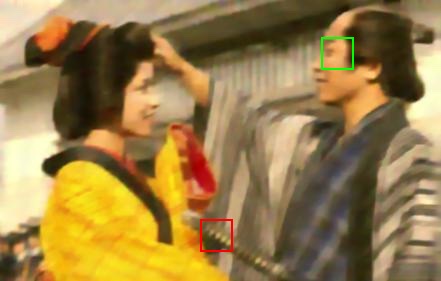}}
& \multicolumn{2}{c}{\includegraphics[width=0.15\textwidth]{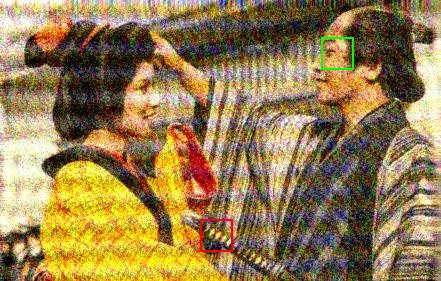}}
&
\multicolumn{2}{c}{\includegraphics[width=0.15\textwidth]{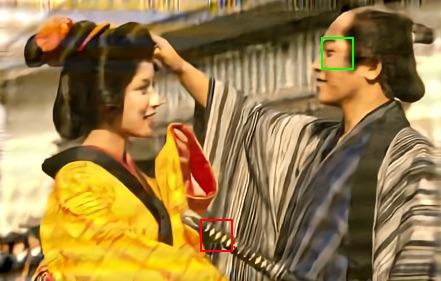}}
& \multicolumn{2}{c}{\includegraphics[width=0.15\textwidth]{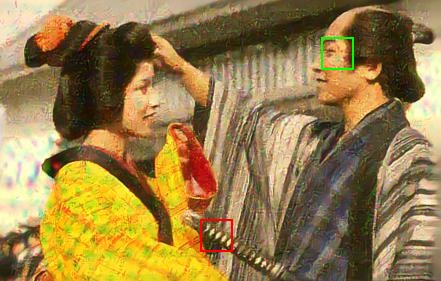}}
& \multicolumn{2}{c}{\includegraphics[width=0.15\textwidth]{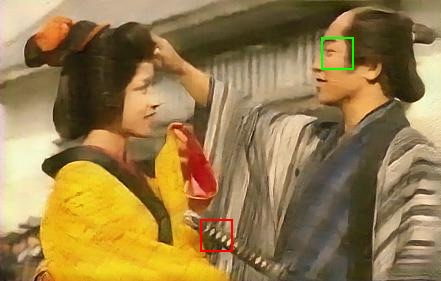}}
\\
\includegraphics[width=0.075\textwidth]{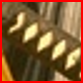}&
\includegraphics[width=0.075\textwidth]{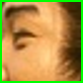}&
\includegraphics[width=0.075\textwidth]{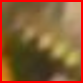}&
\includegraphics[width=0.075\textwidth]{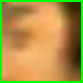}&
\includegraphics[width=0.075\textwidth]{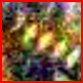}&
\includegraphics[width=0.075\textwidth]{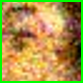}&
\includegraphics[width=0.075\textwidth]{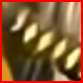}&
\includegraphics[width=0.075\textwidth]{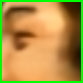}&
\includegraphics[width=0.075\textwidth]{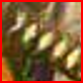}&
\includegraphics[width=0.075\textwidth]{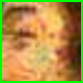}&
\includegraphics[width=0.075\textwidth]{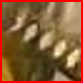}&
\includegraphics[width=0.075\textwidth]{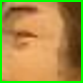}
\\
\multicolumn{2}{c}{Ground Truth}&
\multicolumn{2}{c}{Levin \etal \cite{levin2007image}}&
\multicolumn{2}{c}{CSF \cite{schmidt2014csf}}&
\multicolumn{2}{c}{IRCNN\cite{zhang2017learning}} &
\multicolumn{2}{c}{FDN \cite{kruse2017FFT}}
&
\multicolumn{2}{c}{RGDN (ours)}
\end{tabularx}
\vspace{-0.2cm}
\caption{Visual comparison on the images with different noise levels. The first two rows show results from the dataset \cite{sun2013edge} with $\sigma=2\%$. The bottom two rows show results on an image from the generated BSD-Blur dataset with $\sigma=3\%$.}
\label{fig:syn_res}
\end{center}
\end{figure*}

\subsection{Deconvolution Using the Learned Optimizer}
\label{sec:deonv}
We observe that the proposed optimizer can obtain sustained image quality gain after running with more iterations in testing (as shown in the experiments in Section \ref{sec:exp}), although it is trained using a limited number of steps. Thus we apply the learned optimizer for non-blind deconvolution by using arbitrary steps and stop the processing relying on some certain stopping conditions, similar to a classic optimizer.

\par
{Note that the proposed model shares parameters across different steps. 
By focusing on every single step during training, we can see that the universal optimizer (\ie the GDU) is trained to refine different intermediate images. The optimizer thus can see various images with different status and noise levels. 
The recursive supervisions on all steps encourage the optimizer to lift the image quality of the diverse images in each step. Thus even training with a limited number of steps can render strong generalization for running more steps. 
Moreover, the image prior learned in the first steps is universal to be applied in more stages for image restoration. 
}

\par
{Fig. \ref{fig:inter} shows that the intermediate image qualities are progressively lifted during iterations. Benefited from the learned updating process and the implicit image prior, details are gradually recovered, and the artifacts are suppressed with increased iterations. Considering that the optimizer is trained to improve the quality of each intermediate estimate, the learned optimizer is flexible to afford many iterations to handle the high-level degenerations and fewer iterations for mild degenerations.
}
Thus the learned optimizer is able to generally handle the varying visual appearance among the input images and the intermediate results as well as consistently improve the estimates.

More numerical studies are in Section \ref{sec:study_rec_intest}.
The optimization can be stopped when achieving $|\phi(\bx^t)-\phi(\bx^{t-1})|/|\phi(\bx^t)-\phi(\bx^0)|<\epsilon$, where $\phi(\bx) = \|\by-\bA\bx\|_2^2$ and $\epsilon$ is a small tolerance parameter. In practice, a maximum iteration number $T$ is also used as a stopping criterion.

\section{Experiments}
\label{sec:exp}
We conduct experiments with the proposed method for single image non-blind deconvolution. Our implementation is based on PyTorch \cite{pytorch} and an NVIDIA TITAN X GPU for acceleration. {The code is released at \url{https://github.com/donggong1/learn-optimizer-rgdn}.} In all experiments, the proposed method is evaluated with the \emph{noise-blind} setting. The most compared methods require to know the noise level as input hyper-parameters.

\subsection{Datasets and Experimental Settings}
\label{sec:exp_setting}
\subsubsection{\textbf{Training}}
To generate the triplet set $\{(\bx_i, \by_i, \bk_i)\}_{i=1}^N$ for training, we crop 40,960 RGB images of $256\times 256$ pixels from the PASCAL VOC dataset \cite{Everingham15} as the ground truth images $\bx_i$. 
We then generate the degenerations based on a standard protocol similar to \cite{kruse2017FFT}. 
Specifically, we independently generate $5$ blur kernels according to \cite{chakrabarti2016neural} for each $\bx_i$ and generate blurred image $\by_i$ based on model in Eq. \eqref{eq:blur_model}, which gives 204,800 triplets in total.
After adding a Gaussian noise term from $\cN(\0, \sigma^2\bI )$, 8-bit quantization is used following \cite{kruse2017FFT}. Instead of training a customized model for a specific blur kernel \cite{schuler2013mlp} or noise level \cite{schmidt2014csf,schmidt2016cascades,zhang2016FCN}, we uniformly sample kernel sizes from a set $\{11,21,31,41\}$ and noise levels from an interval $[0.3\%, 1.5\%]$ \footnote{An image $\by$ with a ratio $\sigma$ of Gaussian noise is generated by adding noise from $\cN(\0, \sigma^2\bI)$ for image $\bk*\bx$ with $[0, 1]$ intensity range. $\bI$ denotes the identity matrix.}, which helps to evaluate the ability of the network to handle diverse data. 
{In Section \ref{sec:abl}, we conduct an experiment with 10\% training samples to study the influence of the dataset size on the proposed approach. }

\begin{table*}[!t]
\centering
\caption{Comparison on Levin \etal's dataset \cite{levin2009und}. *Note that the scores of TNRD \cite{chen2016trainable} and GradNet \cite{jin2017noiseblind} are quoted from \cite{jin2017noiseblind}.}
\vspace{-0.2cm}
\label{tab:exp_gray}
\begin{tabular}{cccccccccccc}
\hline
   ~~~$\sigma$~~~      &  ~Mea.~ & ~FD~ & ~Levin~ & ~EPLL~ & ~MLP~ & ~CSF~ & TNRD$^*$ & ~IRCNN~ & GradNet$^*$\! & ~FDN~ & ~RGDN~ \\ \hline
\multirow{2}{*}{~0.59\%~} & PSNR & 32.36  &  33.60 & 34.35 & 31.55  & 32.08 & -- & 33.35 & -- & \bf36.15 & 35.04  \\ \cline{2-12}
                   & SSIM & 0.917  &  0.934 & 0.941 & 0.876 & 0.916 & -- & 0.884 & -- & \bf0.965 & 0.954  \\ \hline
\multirow{2}{*}{1\%} & PSNR & 30.85  & 32.01  & 32.45 & 30.68  & 28.12 & 28.88 & 33.14 & 31.43 & 33.62 &  \bf 33.68 \\ \cline{2-12}
                   & SSIM & 0.892  & 0.913  & 0.930 & 0.882  & 0.828 & 0.854 & 0.896 & 0.912 & 0.949 & \bf 0.954  \\ \hline
\multirow{2}{*}{2\%} & PSNR & 28.84  &  29.92 & 30.03 & 28.16  & 21.68 & 28.10 & 30.09 & 28.88 & 29.70 & \bf 31.01  \\ \cline{2-12}
                   & SSIM & 0.851  & 0.877  & 0.883 & 0.841  & 0.594 & 0.824 & 0.887 & 0.841 & 0.896 & \bf 0.899  \\ \hline
\end{tabular}
\end{table*}

\begin{table*}[]
\centering
\caption{Comparison on 640 RGB images from \cite{sun2013edge} and \cite{sun2012super}. *Note that the scores of EPLL \cite{zoran2011epll}, TNRD \cite{chen2016trainable} and GradNet \cite{jin2017noiseblind} are quoted from \cite{jin2017noiseblind} as a reference.}
\vspace{-0.2cm}
\label{tab:exp_rgb}
\begin{tabular}{cccccccccccc}
\hline
   ~~~$\sigma$~~~      &  ~Mea.~ & ~FD~ & ~Levin~ & ~EPLL$^*$ & ~MLP~ & ~CSF~ & TNRD$^*$ & ~IRCNN& GradNet$^*$\! & ~FDN~ & ~RGDN~ \\ \hline
\multirow{2}{*}{1\%}   & PSNR & 29.90  &  30.29  & 32.05 & 31.01  & 28.32   & 30.03 & 30.44 & 31.75 & \bf 32.52 & 32.33  \\ \cline{2-12}
                       & SSIM & 0.826  &  0.841  & 0.880 & 0.882  & 0.797   & 0.844 & 0.900 & 0.873 &  \bf 0.909 & 0.907  \\ \hline
\multirow{2}{*}{2\%}   & PSNR & 29.08  &  28.81  & 29.60 & 27.82  & 20.06   & 28.79 & 29.47 & 29.31 & 29.04 & \bf 29.59 \\ \cline{2-12}
                       & SSIM & 0.816  &  0.795  & 0.807 & 0.789  & 0.362   & 0.790 & \bf 0.867 & 0.798 & 0.842 &  0.855  \\ \hline
\multirow{2}{*}{3\%}   & PSNR & 23.19  &  28.00  & 28.25 & 25.30  & 16.66   & 28.04 & 28.05 & 28.04 & 24.41 & \bf 28.45  \\ \cline{2-12}
                       & SSIM & 0.532  &  0.768  & 0.758 & 0.627  & 0.237   & 0.750 &  0.806 & 0.750 & 0.653 & \bf 0.812  \\ \hline
\end{tabular}
\end{table*}

\subsubsection{\textbf{Testing}}
The testing is performed on several benchmark datasets \cite{levin2009und,sun2013edge,bsd500} that are independent to the training data.
Considering that RGB images are predominant in reality, we trained our model on RGB images with 3 channels. To test on the benchmark dataset \cite{levin2009und} of gray images, we replicate the single existing channel twice.
Different noise levels are used to measure the robustness of the methods.
In the experiments, we apply the stopping conditions introduced in Section \ref{sec:deonv} and set the maximum iteration number as $T=30$ if not indicated otherwise.

\par
In the following, we will first conduct a full numerical comparison with other state-of-the-art methods, \eg FD \cite{krishnan2009fast}, the method of Levin \etal \cite{levin2007image}, EPLL \cite{zoran2011epll}, MLP \cite{schuler2013mlp}, CSF \cite{schmidt2014csf}, TNRD \cite{chen2016trainable}, IRCNN \cite{zhang2017learning} and FDN \cite{kruse2017FFT}. 
{The conventional optimization based methods usually rely on some empirically designed priors/regularizers, \eg sparse gradient prior \cite{krishnan2009fast,levin2009und} and the GMM prior \cite{zoran2011epll}. 
These methods optimize the problem relying on some advanced variants (\eg some splitting technique based algorithms) of the standard gradient descent algorithm for efficiency \cite{chambolle2010introduction}. For example, the methods in \cite{krishnan2009fast,zhang2016FCN} are based on the half-quadratic splitting based optimization algorithms.
}
We then conduct a series of empirical analyses and ablation studies for the proposed method. Finally, quantitative comparisons between the methods are conducted on real-world images.
It is worth noting that, in deconvolution, apart from the RGDN that is \emph{free of parameters}, the parameters of all other methods are set using the ground truth noise level.
We use the pairwise version of CSF \cite{schmidt2014csf} trained for deconvolution in comparison.
The comparison with the CNN based baseline method \cite{xu2014deep} is absent since it needs fine-tuning for every blur kernel, making it unpractical.
We measure the performance in terms of PSNR and SSIM \cite{wang2004ssim}. Following \cite{kruse2017FFT}, the regions close to the image boundary are discarded when calculating the measurements.

\subsection{Numerical Evaluations on Synthetic Datasets}
\label{sec:num_eval_syn}

\subsubsection{\textbf{Evaluation on grayscale image benchmark}}
We first evaluate the performance of the methods on a widely used benchmark dataset of Levin \etal \cite{levin2009und}, which contains 32 blurry gray images (of $255\times 255$ pixels) from 4 clear images and 8 blur kernels. To deal with the gray images, we generate 3-channel images via replication. Images with different noise levels ($\sigma=0.59\%, 1\%, 2\%$) are also generated by adding adding noise to all channels. Note that the noise level on the original blurry images are about $0.59\%$, as discussed in \cite{kruse2017FFT}.
The comparison on the three noise levels is shown in Table \ref{tab:exp_gray}. IRCNN \cite{zhang2017learning}, FDN \cite{kruse2017FFT} and RGDN outperform other methods due to the deep neural networks that provide more powerful natural image priors.
Although RGDN is trained as a noise-level-versatile model, its performance is better than other methods or competitive with the best one.
The performance of EPLL \cite{zoran2011epll} is close to the best one on this benchmark, however it is dozens of times slower than the proposed methods.

\subsubsection{\textbf{Evaluation on large RGB images}}
We evaluate the methods on the dataset \cite{sun2013edge} with large images.
We generate an RGB version of the benchmark \cite{sun2013edge} using the original 80 RGB images \cite{sun2012super} and same 8 blur kernels from Levin \etal's dataset \cite{levin2009und}. Three different noise levels are adopted.
The average PSNR and SSIM values are shown in Table \ref{tab:exp_rgb}.
The performance of RGDN is on par with or better than other methods. Maybe because FDN \cite{kruse2017FFT} takes the ground truth noise level as input, it achieves marginally better performance than the proposed method when the noise level is low ($\sigma=1\%$). Even though the RGDN is trained on the data with noise level lower than $1.5\%$, it still performs well on high noise level data (\ie $\sigma=2\%$ and $3\%$), which also proves the generalization ability of the proposed method.
IRCNN \cite{zhang2017learning} also performs very well for large noise levels.
An example of visual comparison is shown in Fig. \ref{fig:syn_res}.

\subsubsection{\textbf{Evaluation on images with large blur kernels and strong noise}}
The above datasets only use 8 blur kernels from \cite{levin2009und}, whose sizes are limited to $27\times 27$. To study the behaviors of the methods on large blur kernels, we generate a dataset (BSD-Blur) with 150 images by randomly selecting 15 images from the dataset BSD \cite{bsd500} and 10 blur kernels of size $41\times 41$ from \cite{chakrabarti2016neural}. In order to study the noise robustness, high noise levels (2\%, 3\% and 5\%) are used.
As shown in Table \ref{tab:exp_bsd}, IRCNN \cite{zhang2017learning} and the proposed method significantly outperform the other methods. However, as shown in Fig. \ref{fig:syn_res}, the results of IRCNN \cite{zhang2017learning} suffer from more ringing artifacts and over-smoothness, which may be related to the conventional HQS image updating scheme in IRCNN.
The performance of FDN \cite{kruse2017FFT} degenerated quickly with increasing of noise level, although it is trained on a dataset with a similar noise level setting to ours.
The proposed method achieves better generalization on the testing data.
As shown in Fig. \ref{fig:syn_res}, the visual quality of the image recovered by the proposed method also outperforms the other methods. Even the input image is degenerated by severe noise, the proposed method can still recover a clear image with rich details.

\begin{table*}[]
\centering
\caption{Comparison on 150 images from BSD-Blur with larger blur kernel and strong noise.}
\vspace{-0.2cm}
\label{tab:exp_bsd}
\begin{tabular}{ccccccccc}
\hline
   ~~~$\sigma$ ~~~      &  ~~Mea.~~~~ & ~FD~ & ~Levin~  & ~MLP~ & ~CSF~ & IRCNN & ~FDN~ & ~RGDN~ \\ \hline
\multirow{2}{*}{2\%} & PSNR & 23.60  &  22.70  & 19.23 & 17.40  &  22.29 & 23.48 & \bf 24.27  \\ \cline{2-9}
                   & SSIM & 0.648  &  0.577  & 0.570 & 0.497 & 0.657 & 0.697  & \bf 0.699  \\ \hline
\multirow{2}{*}{3\%} & PSNR & 20.65  & 22.12   & 19.71  &  15.15  & 22.03 & 20.25 &  \bf 23.17 \\ \cline{2-9}
                   & SSIM & 0.555 & 0.541   & 0.546  & 0.385 & 0.654 & 0.559 & \bf 0.637  \\ \hline
\multirow{2}{*}{5\%} & PSNR & 6.410  &  21.48  & 19.87  & 12.51  & 21.10 & 9.090 & \bf 21.80  \\ \cline{2-9}
                   & SSIM & 0.004  & 0.501    & 0.500  & 0.259 & \bf 0.605 & 0.094 &  0.560  \\ \hline
\end{tabular}
\end{table*}

\begin{table*}[htp]
\centering
{
\caption{Comparison on Levin \etal's dataset \cite{levin2009und} with estimated blur kernels. The blur kernels are estimated using \cite{gong2016blind}. }
\vspace{-0.2cm}
\label{tab:exp_gray_estk}
\begin{tabular}{cccccccccc}
\hline
   ~~~$\sigma$~~~      &  ~Mea.~ & ~FD~ & ~Levin~ & ~EPLL~ & ~MLP~ & ~CSF~ & ~IRCNN~ & ~FDN~ & ~RGDN~ \\ \hline
\multirow{2}{*}{~0.59\%~} & PSNR & 29.48  & 30.17 & 30.31 & 27.80  & 29.41 & 27.76 & 29.35  & \bf 30.51  \\ \cline{2-10}
                          & SSIM & 0.889  & 0.905 & 0.923 & 0.838  & 0.892 & 0.826 & 0.914 & \bf 0.922  \\ \hline
\multirow{2}{*}{1\%}      & PSNR & 28.80  & 29.52 & 30.12 & 27.70  & 27.51 & 28.36 & 29.50 & \bf 30.16 \\ \cline{2-10}
                          & SSIM & 0.868  & 0.888 & 0.912 & 0.843  & 0.831 & 0.846 & 0.914  & \bf0.911  \\ \hline
\multirow{2}{*}{2\%}      & PSNR & 27.61  & 28.38 & 28.98 & 26.72  & 22.63 & 28.54 & 28.72  & \bf 29.10  \\ \cline{2-10}
                          & SSIM & 0.832  & 0.857 & 0.875 & 0.812  & 0.632 & 0.848 & 0.892  & \bf 0.880  \\ \hline
\end{tabular}
}
\end{table*}

{
\subsubsection{\textbf{Experiments on the estimated blur kernels}}
Non-blind deconvolution is often applied as a subcomponent of blind deblurring, where the blur kernels are estimated by other methods \cite{pan2014text,gong2016blind} and thus not completely accurate. 
To evaluate the robustness to kernel estimation error, we conduct experiments based on Levin \etal's \cite{levin2009und} images and the blur kernels estimated by \cite{gong2016blind}. 
Considering that image noise usually severely influences kernel estimation, we apply the kernels estimated on Levin \etal's original dataset (with noise level as $0.59\%$) for non-blind deconvolution on images with different noise levels. 
The kernels contain mild estimation errors and are propitious to evaluating the deconvolution methods. The results in Table \ref{tab:exp_gray_estk} show that the proposed method is more robust to kernel estimation error. Results of RGDN suffer less from artifacts caused by kernel error. Note that all the learning-based methods are trained with accurate blur kernels. The visual results on real-world images in Section \ref{sec:exp_real_vis} show the superiority of the proposed method further. 
}

\subsection{Ablation Study}
\label{sec:abl}
In this section, we perform an ablation study to analyze several aspects in terms of the structure of the RGDN. For simplicity, we run all the studies on the Levin \etal's dataset \cite{levin2009und} with different noise levels used in Section \ref{sec:num_eval_syn}.

\subsubsection{\textbf{Study on the structure of RGDN}}
As shown in Fig. \ref{fig:rgdn_overview}, RGDN mainly consists of three subnetworks corresponding to three parameterized operations $\cR(\cdot)$, $\cH(\cdot)$ and $\cD(\cdot)$, which are trained jointly.
To verify the importance of each subnetwork, we conduct experiments by removing them from RGDN, respectively, and train the networks with the same setting for the complete RGDN.
Table \ref{tab:ablation} shows that removing the regularization term $\cR(\cdot)$ substantially degrades results, showing that $\cR(\cdot)$ is crucial for RGDN. The RGDN without $\cR(\cdot)$ corresponds to the problem \eqref{eq:devon_min_prob} without the regularizer $\Omega(\bx)$, which suffers from the ill-posedness. 
{We also studied the model variants by removing $\cR(\cdot)$ and adding more layers into $\cH(\cdot)$ and $\cD(\cdot)$, respectively, which match the model capacity of the full model RGDN. We report the results in Table \ref{tab:ablation} as ``w/o $\cR(\cdot)$ w/ $\cH^+(\cdot)$'' and w/o $\cR(\cdot)$, w/ $\cD^+(\cdot)$. }
Removing both the direction scaling operator $\cD(\cdot)$ and $\cH(\cdot)$ also significantly degrades the performances, {even after adding a large $\cR^+(\cdot)$}. This may be interpreted as the deficiency of the ability to handle noise. 
Table \ref{tab:ablation} also shows that the performance degenerates without $\cD(\cdot)$ or $\cH(\cdot)$, and the direction scaling operator $\cD(\cdot)$ plays a more important role than $\cH(\cdot)$.
The three terms are all important to the results, and work interdependently.

\begin{table}[!t]
\centering
\caption{Ablation study: performances of different structures of our method on Levin \etal's dataset \cite{levin2009und}. }
\label{tab:ablation}
\vspace{-0.2cm}
\begin{tabular}{lcccccc}
\hline
\multirow{2}{*}{} & \multicolumn{2}{c}{$\sigma=0.59\%$} & \multicolumn{2}{c}{$\sigma=1\%$} & \multicolumn{2}{c}{$\sigma=2\%$} \\ \cline{2-7}
                  & PSNR         & SSIM         & PSNR          &   SSIM       &  PSNR         &     SSIM     \\ \hline
\!\!\!w/o $\cD(\cdot)$\!\! &  33.71   &     0.939    &    32.61   &  0.928  & 30.47          &   0.886     \\
\!\!\!w/o $\cH(\cdot)$\!\! &    33.73       &   0.941      &   33.00   &   0.929   &     30.92      &    0.892    \\
\!\!\!w/o $\cD(\cdot)$ and $\cH(\cdot)$\!\!\!\!\!\!\!\!\! &    11.33       &      0.211     &11.32           &     0.210    &     11.30    &    0.202     \\ 
\!\!\!w/ only $\cR^+(\cdot)$\!\! &  12.52   &  0.243  & 12.49   & 0.240  &  12.47     & 0.233       \\  \hline
\!\!\!w/o $\cR(\cdot)$\!\! &  18.94   &  0.629  &  18.69  &  0.619 &     18.44    & 0.588       \\ 
\!\!\!w/o $\cR(\cdot)$ w/ $\cH^+\!(\cdot)$ \!\!\!\!\!\!\!\!\! &  19.40   &  0.702  &  19.60  &  0.698  &   19.81    & 0.673      \\ 
\!\!\!w/o $\cR(\cdot)$ w/ $\cD^+\!(\cdot)$ \!\!\!\!\!\!\!\!\! &  23.19   &  0.865  &  23.22  &  0.846 &    22.71    & 0.786       \\ 
\hline
\!\!\!RGDN &   \bf 35.04        &   \bf 0.954       &     \bf 33.68     &   \bf 0.954        & \bf 31.01         &   \bf 0.899        \\ \hline
\end{tabular}
\end{table}

\subsubsection{\textbf{Study on the recursive supervision}} We use recursive supervision to accelerate training and enable the learned optimizer to push the image towards the ground truth in each step.
{In this section, we study the importance of recursive supervision and the behaviors of the losses on the intermediate steps. We treat the loss items on all steps equally and set $\kappa_t=1, t=1,..., S$ as default in our implementation. }

\par
{We firstly study the importance of the recursive supervision by directly removing the recursive supervision on intermediate steps and only keeping the one on the last step, \ie setting $\kappa_t=0, t=1,..., S-1$ and $ \kappa_S=1$. We show the results of the trained model on the data of Levin \etal in Table \ref{tab:abl_loss_step_weights} and refer the model as ``w/o inter. losses''. 
As shown in Table \ref{tab:abl_loss_step_weights}, removing the recursive supervision incurs a significant performance degeneration due to the difficulties of training. Only imposing supervision on the final step restricts the training merely minimizing the loss after a fixed number of steps, making the leaned optimizer less flexible.}

\par
{We set $\kappa_t=1, t=1,...,S$ as the default in our implementation. 
To further study the behavior of the recursive supervision, we conducted experiments by setting varying weights for the losses in intermediate steps in ascending or descending order. For convenience, we set the weights by letting $\kappa_t=\eta^{S-t}$ in experiments, where $\eta>0$ is a scalar denoting the rate of ascending or descending. We tested two settings of $\eta=1.1$ and $\eta=0.9$, which correspond to two settings for the weights, \ie $(1.4641, 1.331, 1.2100, 1.1000, 1.000)$ and $(0.6561, 0.7290, 0.8100, 0.9000, 1.0000)$. The results in Table \ref{tab:abl_loss_step_weights} show that the supervisions on early steps are important for the performance, and the model can obtain satisfactory results with different settings for the weights. 
There is only a small gap between the two settings, $\eta=1$ and $\eta=0.9$. The default setting $\eta=1$ is a practical and convenient choice in the real application. 
We observe that weights in ascending order can obtain better results than that in descending order, which implies that the supervisions on the later steps are more important than the ones in early steps. 
}

\begin{table}[!t]
\centering
{
\caption{Ablation Study on the weights for the training losses in different steps. The results are evaluated on the datasets from Levin \etal \cite{levin2009und}. 
Supervisions on the intermediate steps apart from the last step are disabled for ``w/o inter. losses''. 
}
\label{tab:abl_loss_step_weights}
\vspace{-0.2cm}
\begin{tabular}{lcccccc}
\hline
\multirow{2}{*}{} & \multicolumn{2}{c}{$\sigma=0.59\%$} & \multicolumn{2}{c}{$\sigma=1\%$} & \multicolumn{2}{c}{$\sigma=2\%$} \\ \cline{2-7}
                  & PSNR         & SSIM         & PSNR          &   SSIM       &  PSNR         &      SSIM     \\ \hline
\!\!w/o inter. losses\!\!\! &  22.04   & 0.744   &  21.54  & 0.709  &  20.82     &   0.655     \\
\!\!Ascending $\kappa_t$\!\!\! &  34.06   & 0.9419   &  32.80  & 0.926  &  30.86     &   0.895     \\
\!\!Descending $\kappa_t$\!\!\! &  33.03   &    0.936   &  31.71   &  0.918  & 29.80      &  0.881   \\
\!\!Ours ($\kappa_t=1$)\!\!\! &   \bf 35.04        &   \bf 0.954       &     \bf 33.68     &   \bf 0.954        & \bf 31.01         &   \bf 0.899        \\ \hline
\end{tabular}
}
\end{table}

\begin{table}[!t]
\centering
{
\caption{Ablation study on the loss function weight $\tau$. $\tau$ is the weight on the loss term on image gradients. }
\label{tab:abl_grad_loss_weight}
\vspace{-0.2cm}
\begin{tabular}{lcccccc}
\hline
\multirow{2}{*}{} & \multicolumn{2}{c}{$\sigma=0.59\%$} & \multicolumn{2}{c}{$\sigma=1\%$} & \multicolumn{2}{c}{$\sigma=2\%$} \\ \cline{2-7}
                  & PSNR         & SSIM         & PSNR          &   SSIM       &  PSNR        &      SSIM     \\ \hline
$\tau=0.5$ & 34.30    &  0.951  &  32.93  & 0.932  &  28.90     &   0.831     \\
$\tau=2$ &  34.73   &     0.953    &  \bf  33.69   &  0.941 & 30.92       &   0.890    \\
$\tau=1$ (default) &   \bf 35.04        &   \bf 0.954       &      33.68     &   \bf 0.954        & \bf 31.01         &   \bf 0.899        \\ \hline
\end{tabular}
}
\end{table}

{
\subsubsection{\textbf{Study on the weights on gradient loss}} We use the loss on image gradients as many previous methods \cite{zhang2016FCN,fan2017generic} and observed that the gradient loss is beneficial to the performance. 
As defined in Eq. \eqref{eq:obj_step_weight}, we use the weight $\tau$ to control the strength of the loss term on image gradients. 
In Table \ref{tab:abl_grad_loss_weight}, we show the results of the model trained with some different settings on $\tau$. 
All models trained with different weights can obtain satisfactory results, although the hyper-parameters (such as learning rate) are not specifically tuned. 
We can observe that the models trained with different (not too small) $\tau$'s are all better or comparative with other methods. 
}

{
\subsubsection{\textbf{Training with a small dataset}} 
We study the influence of the dataset size by training a model using a dataset smaller than the default dataset introduced in Section \ref{sec:exp_setting}. Specifically, we use 10\% of the original ground truth images and the same protocol for generating degenerations to synthesize a 10\%-size training dataset.  
The results are shown in Table \ref{tab:abl_less_data}. 
The performance of the model trained with small dataset decreases a bit, which is comparative with or better than other methods. Same to our standard model, the model trained with small dataset does not need to know the noise levels, which is superior to the compared methods. 
}

\begin{table}[!t]
\centering
{
\caption{Training with a smaller dataset. Evaluation is conducted on Levin \etal's dataset.}
\label{tab:abl_less_data}
\vspace{-0.2cm}
\begin{tabular}{lcccccc}
\hline
\multirow{2}{*}{} & \multicolumn{2}{c}{$\sigma=0.59\%$} & \multicolumn{2}{c}{$\sigma=1\%$} & \multicolumn{2}{c}{$\sigma=2\%$} \\ \cline{2-7}
                  & PSNR         & SSIM         & PSNR          &   SSIM       &  PSNR        &      SSIM     \\ \hline
10\% tr. data & 34.46   &  0.955  &  33.30  &  0.939 &   30.01   &  0.869    \\
Ours &   \bf 35.04        &   \bf 0.954       &     \bf 33.68     &   \bf 0.954        & \bf 31.01         &   \bf 0.899        \\ \hline
\end{tabular}
}
\end{table}

\subsection{Empirical Convergence Analysis}
\label{sec:study_rec_intest}
\begin{figure}[htp]
\centering
\subfigure[Levin \etal \cite{levin2009und}]{
\centering
\includegraphics[trim =-2mm 0mm 0mm 0mm, clip,
width=0.45\linewidth]{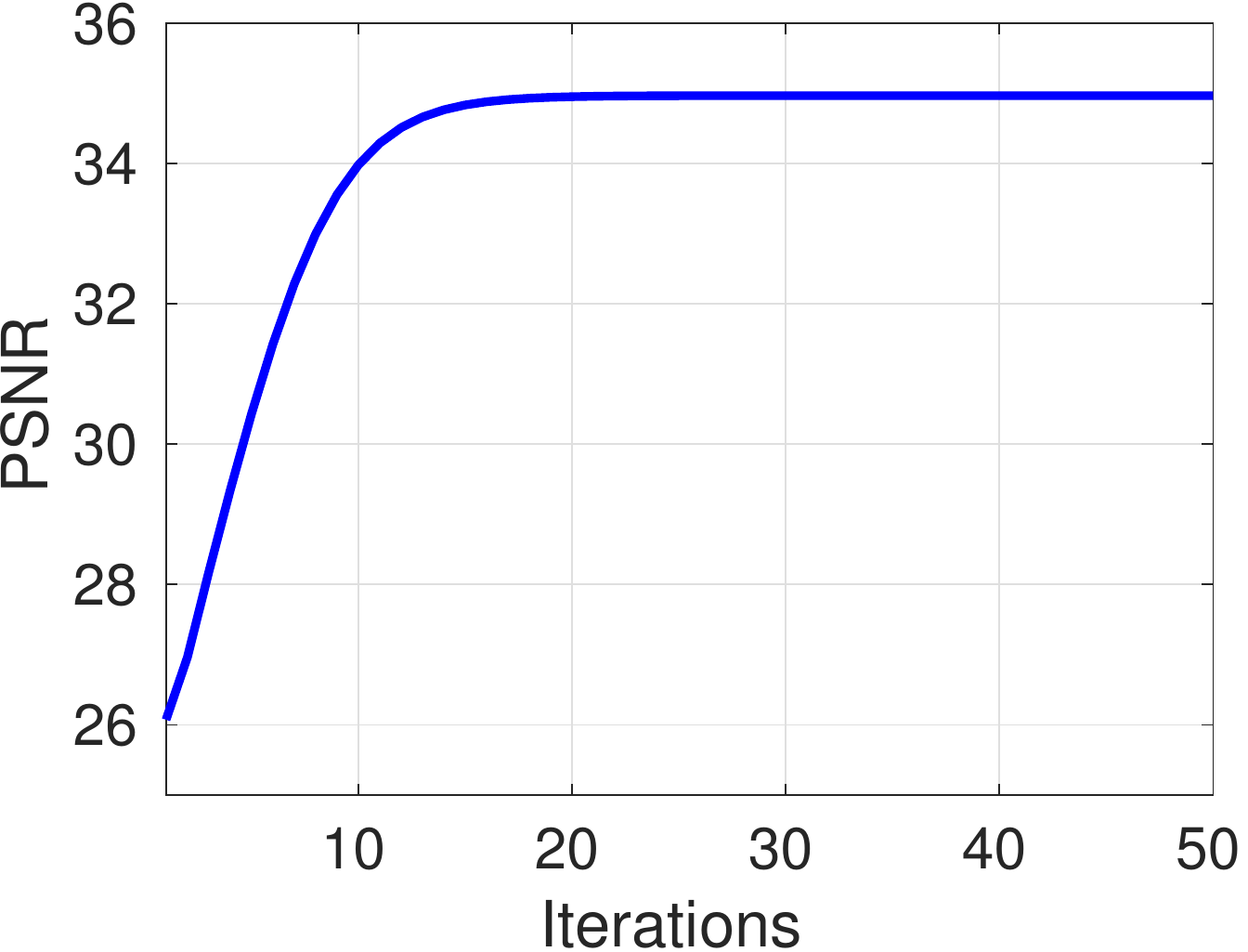}
\includegraphics[trim =0mm 0mm 0mm 0mm, clip, width=0.48\linewidth]{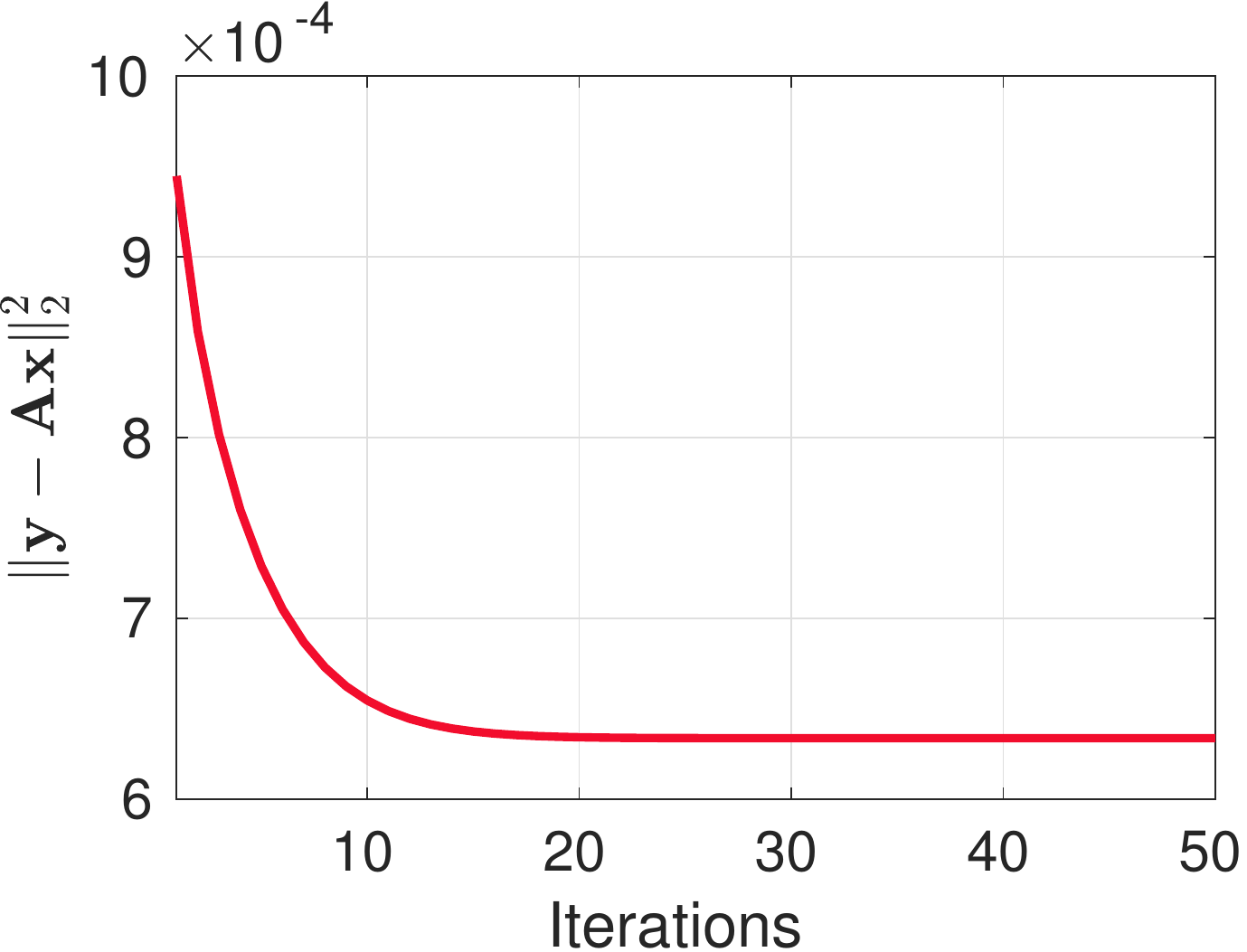}
\label{fig:psnr_vs_ite_levin}
}
\subfigure[BSD-Blur]{
\label{fig:psnr_vs_ite_bsd}
\centering
\includegraphics[trim =-2mm 0mm 0mm 0mm, clip,
width=0.45\linewidth]{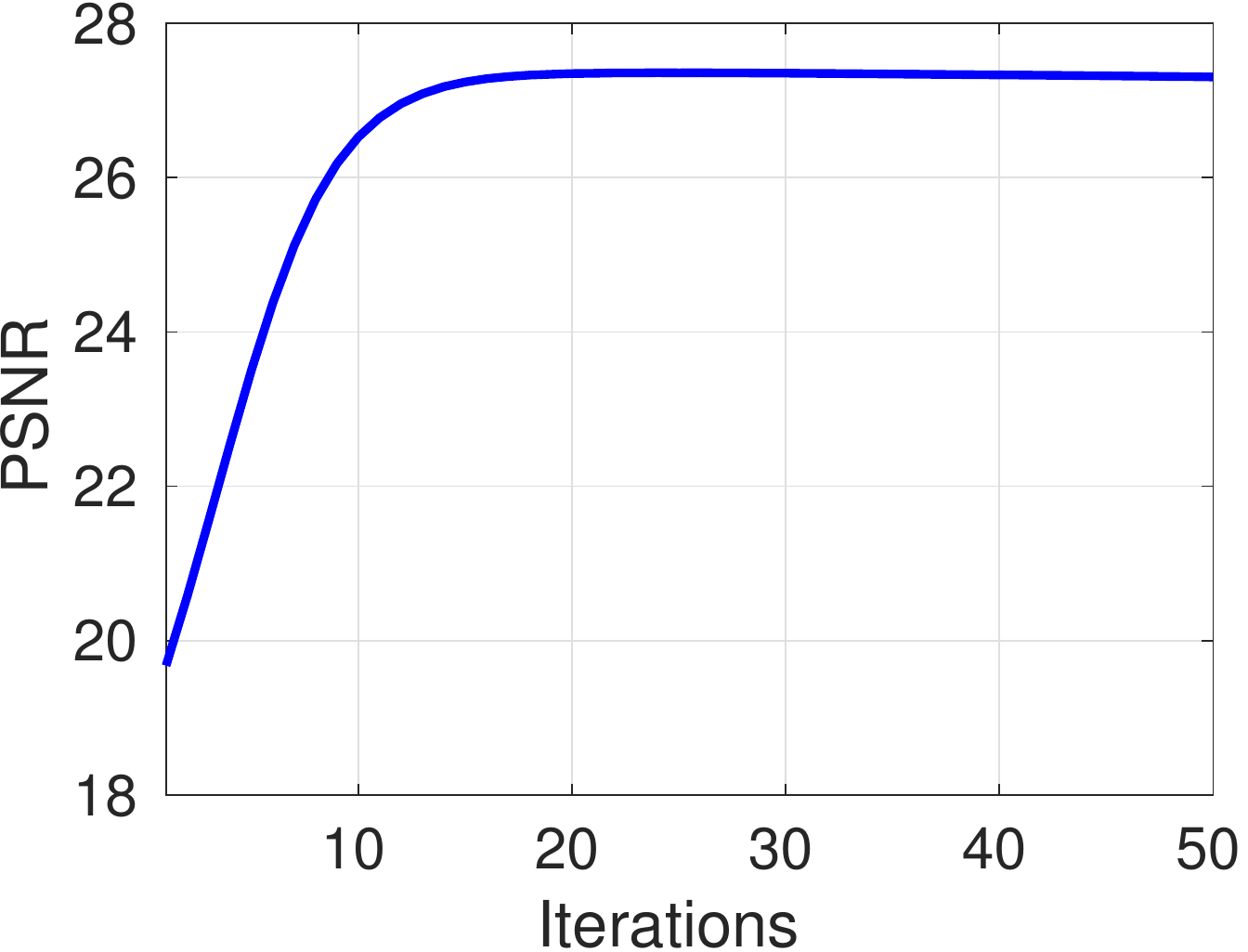}
\includegraphics[trim =0mm 0mm 0mm 0mm, clip,
width=0.47\linewidth]{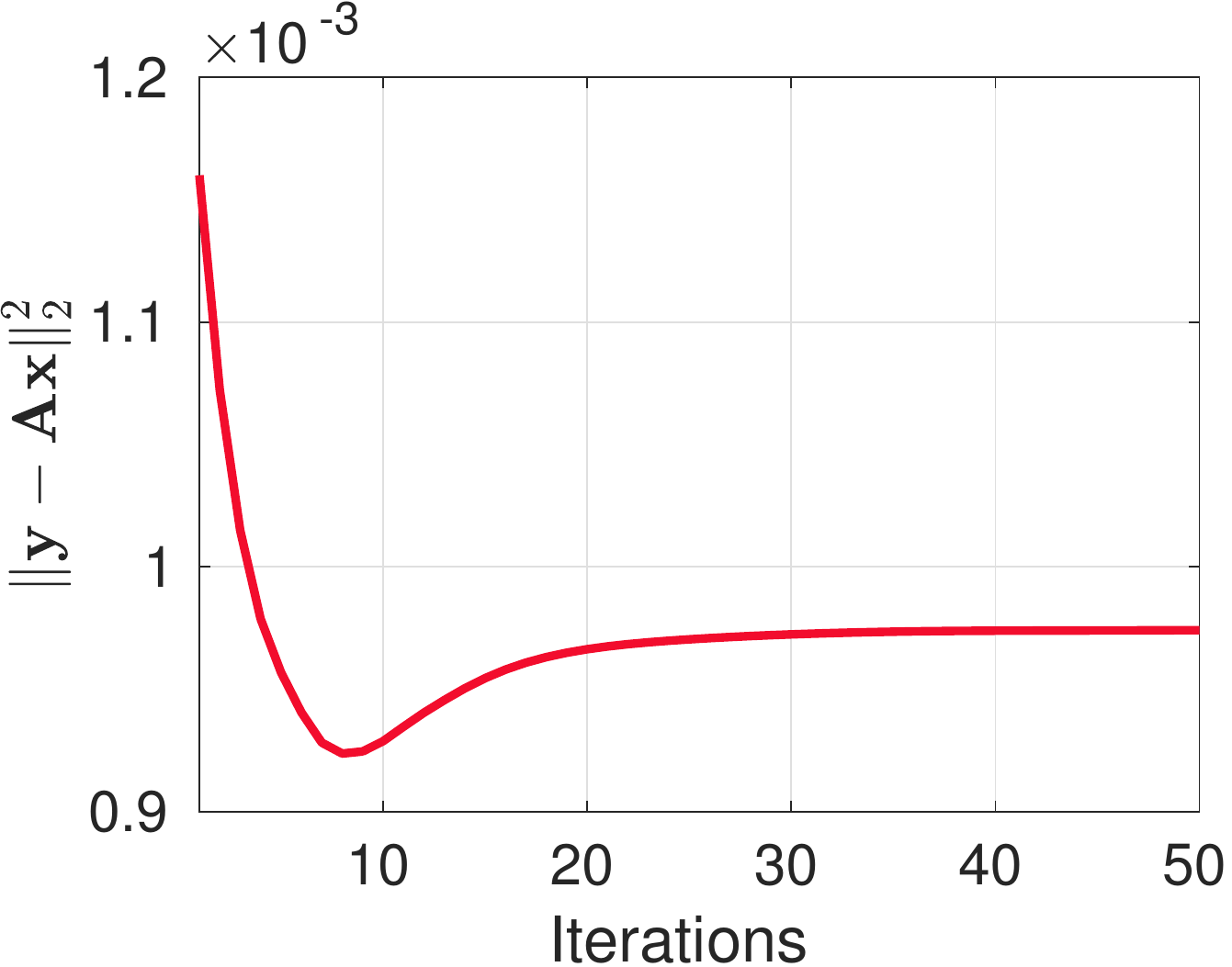}
}
\caption{PSNR / fitting error vs. iteration: empirical convergence analysis of the learned optimizer on image deconvolution.
(a) and (b) are the results on the images from Levin \etal's dataset \cite{levin2009und} and the BSD-Blur, respectively.}
\label{fig:psnr_vs_ite}
\end{figure}

Since the neural networks are too complicated to derive some general properties for convergence analysis, we tend to empirically analyze the convergence of the learned optimizer in testing. To study the instance-specific convergence speed in the meantime, we select two images from Levin \etal's dataset \cite{levin2009und} and BSD-Blur, respectively, and perform deconvolution on them.
Fig. \ref{fig:psnr_vs_ite} shows the variation of the PSNR and the data fitting error $\|\by-\bA\bx\|_2^2$ with increasing iteration numbers. As more updating steps are performed, PSNR values smoothly increase and the fitting error decrease, which is consistent with results shown in Fig. \ref{fig:inter}.
The empirical results in Fig. \ref{fig:psnr_vs_ite} show that the learned optimizer converges well after 30 iterations.
Moreover, Fig. \ref{fig:psnr_vs_ite} also shows that different images require different numbers of steps for convergence.
Comparing the the previous methods with a fixed number of steps \cite{kruse2017FFT,schmidt2014csf,zhang2016FCN}, the learned universal optimizer provides a flexibility to fit the different requirements for different images.

\subsection{Visual Comparison on Real-world Images}
\label{sec:exp_real_vis}
In real-world applications, the non-blind deconvolution performs as a part of the blind deblurring \cite{pan2014text,gong2017self}, where the ground truth blur kernel is unknown.
The non-blind deconvolution is performed using some imprecise kernels estimated by other methods, \eg \cite{pan2014text,gong2016blind}, which brings more challenges.
We thus conduct experiments to study the practicability of the proposed method.
Since the ground truth images are also unknown, we only present the visual comparison against the state-of-the-art methods.

\par
\begin{figure*}[htp]
\centering
\setlength{\tabcolsep}{1pt}
\hspace{-0.5cm}
\begin{tabularx}{0.96\textwidth}{cccccccc}
\multicolumn{2}{c}{\begin{overpic}[width=0.24\textwidth,trim=2cm 2cm 2cm 2cm, clip]{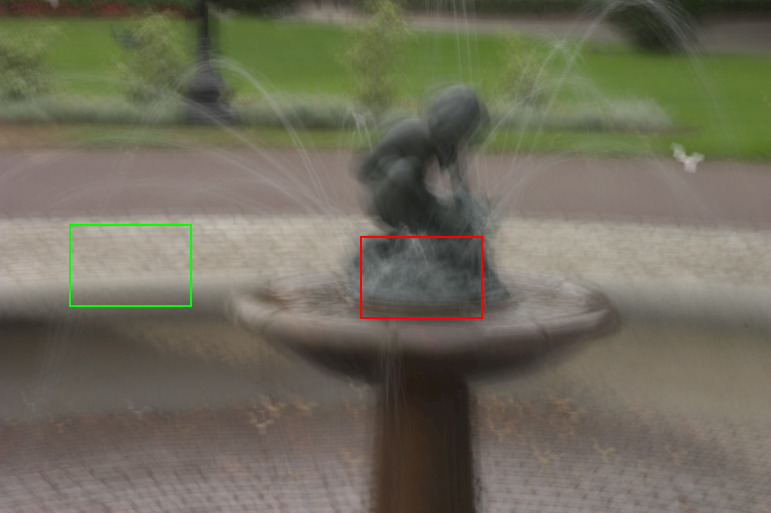}
\put(81, 34){\includegraphics[width=0.08\textwidth]{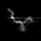}}
\put(2,3){\footnotesize\color{white}{\bf (a) Input}}
\end{overpic}}
&
\multicolumn{2}{c}{\begin{overpic}[width=0.24\textwidth,trim=2cm 2cm 2cm 2cm,, clip]{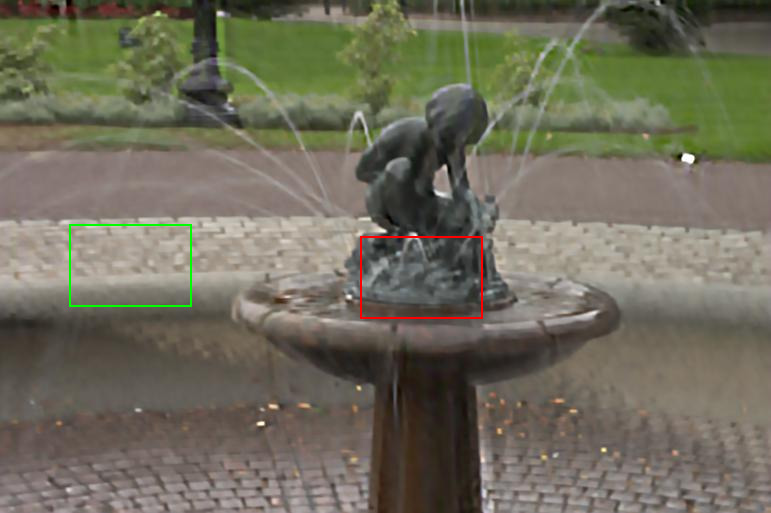}
\put(2,3){\footnotesize\color{white}{\bf (b) Levin \etal \cite{levin2007image}}}
\end{overpic}}
&
\multicolumn{2}{c}{\begin{overpic}[width=0.24\textwidth,trim=2cm 2cm 2cm 2cm,, clip]{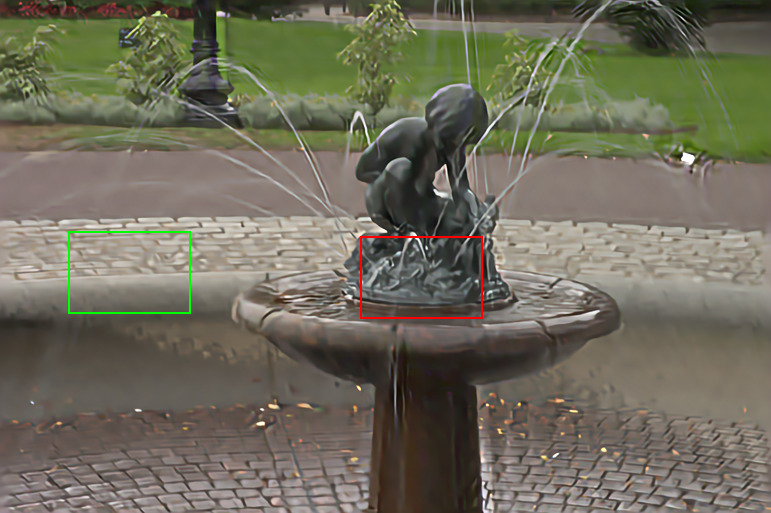}
\put(2,3){\footnotesize\color{white}{\bf (c) CSF \cite{schmidt2014csf} }}
\end{overpic}}
&
\multicolumn{2}{c}{\begin{overpic}[width=0.24\textwidth,trim=2cm 2cm 2cm 2cm, clip]{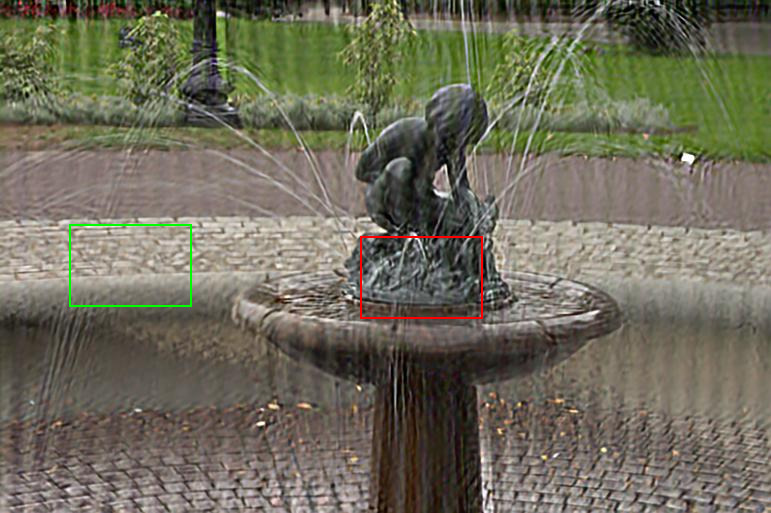}
\put(2,3){\footnotesize\color{white}{\bf (d) MLP \cite{schuler2013mlp} }}
\end{overpic}}\\
\includegraphics[width=0.12\textwidth]{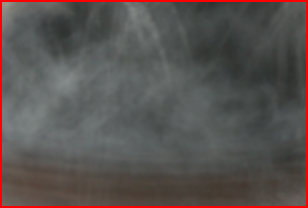}&
\includegraphics[width=0.12\textwidth]{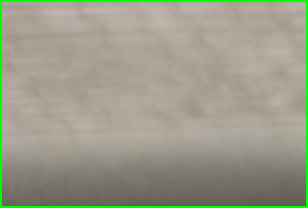}&
\includegraphics[width=0.12\textwidth]{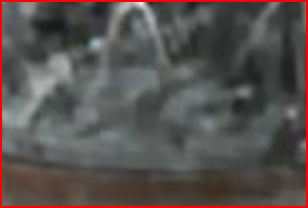}&
\includegraphics[width=0.12\textwidth]{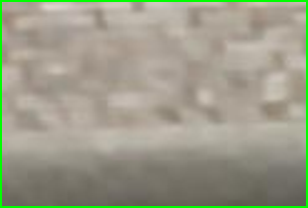}&
\includegraphics[width=0.12\textwidth]{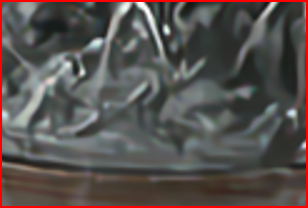}&
\includegraphics[width=0.12\textwidth]{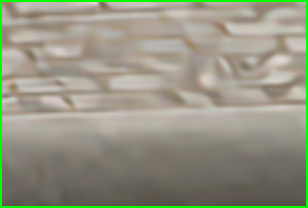}&
\includegraphics[width=0.12\textwidth]{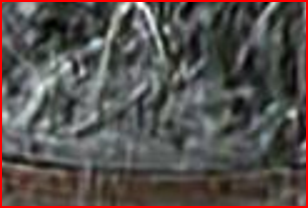}&
\includegraphics[width=0.12\textwidth]{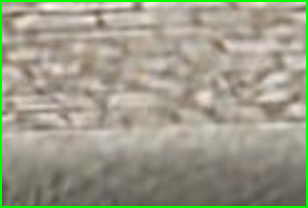}\\
\multicolumn{2}{c}{\begin{overpic}[width=0.24\textwidth,trim=2cm 2cm 2cm 2cm, clip]{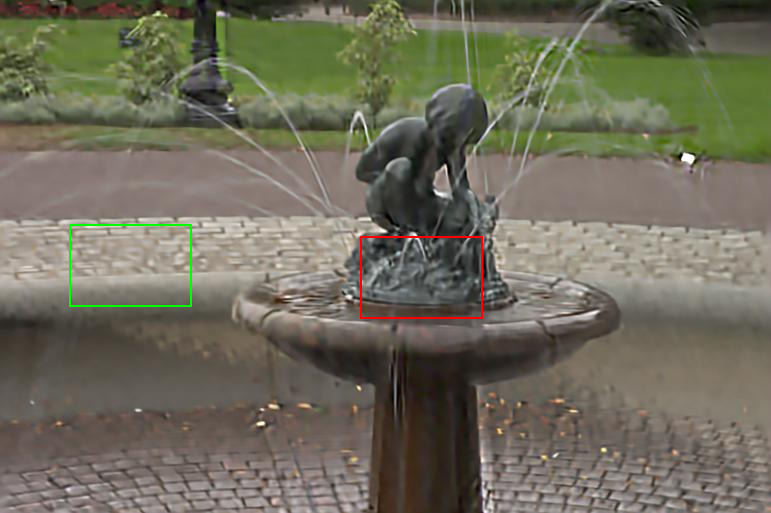}
\put(2,3){\footnotesize\color{white}{\bf (e) EPLL \cite{zoran2011epll}}}
\end{overpic}}
&
\multicolumn{2}{c}{\begin{overpic}[width=0.24\textwidth,trim=2cm 2cm 2cm 2cm,, clip]{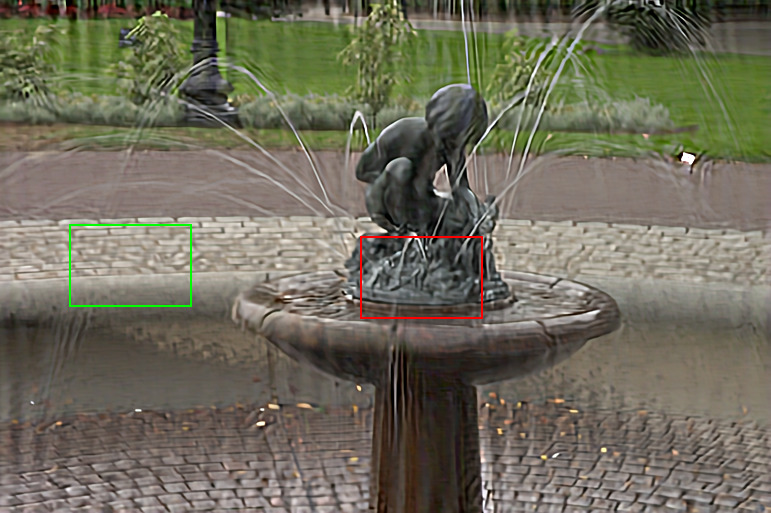}
\put(2,3){\footnotesize\color{white}{\bf (f) IRCNN \cite{zhang2017learning}}}
\end{overpic}}
&
\multicolumn{2}{c}{\begin{overpic}[width=0.24\textwidth,trim=2cm 2cm 2cm 2cm,, clip]{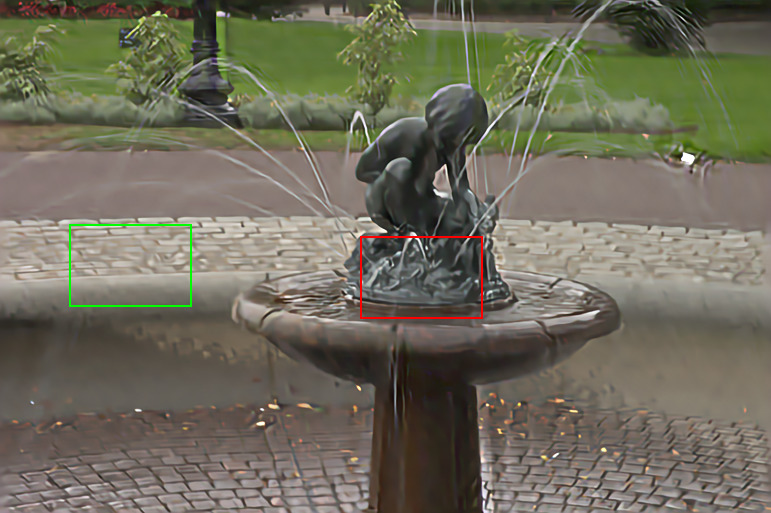}
\put(2,3){\footnotesize\color{white}{\bf (g) FDN \cite{kruse2017FFT}}}
\end{overpic}}
&
\multicolumn{2}{c}{\begin{overpic}[width=0.24\textwidth,trim=2cm 2cm 2cm 2cm, clip]{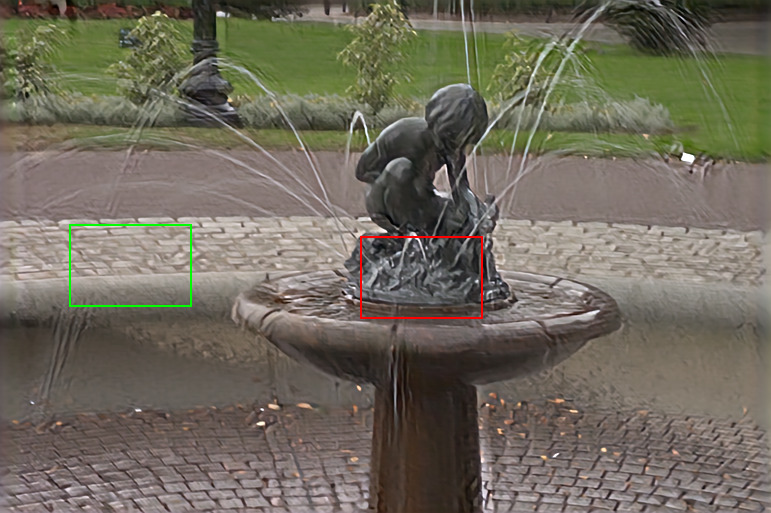}
\put(2,3){\footnotesize\color{white}{\bf (h) RGDN (ours) }}
\end{overpic}}\\
\includegraphics[width=0.12\textwidth]{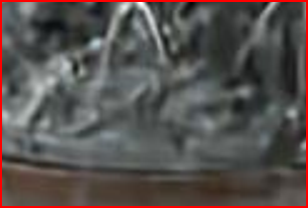}&
\includegraphics[width=0.12\textwidth]{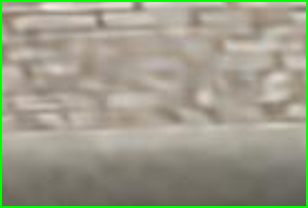}&
\includegraphics[width=0.12\textwidth]{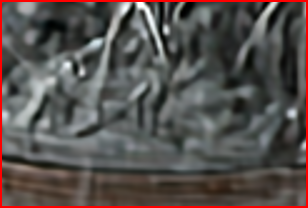}&
\includegraphics[width=0.12\textwidth]{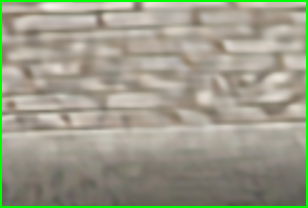}&
\includegraphics[width=0.12\textwidth]{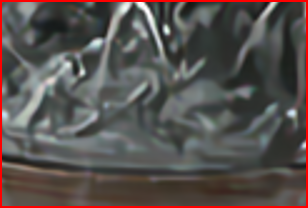}&
\includegraphics[width=0.12\textwidth]{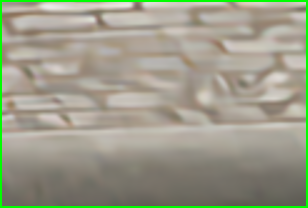}&
\includegraphics[width=0.12\textwidth]{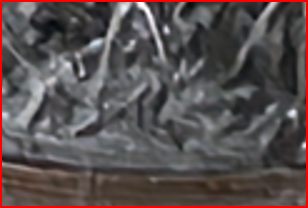}&
\includegraphics[width=0.12\textwidth]{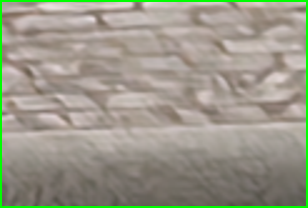}
\end{tabularx}
  \vspace{-0.2cm}
\caption{Deconvolution results on a real-world image. }
  \label{fig:foun}
\end{figure*}

We first test on a real-world image given a blur kernel estimated by \cite{pan2014text}.
As shown in Fig. \ref{fig:foun}, even though the input kernel is imprecise, the proposed method can recover the details of the blurry image and suppress the ringing artifacts due to the powerful learned optimizer. However, the results of other methods suffers from artifacts or over-smoothness due to the inaccurate kernel and the unknown noise level, which also shows that the proposed method is generally more practical in real-world scenarios.

\par
\begin{figure}[!t]
\vspace{-0.1cm}
\centering
\centering
\begin{minipage}[b]{.115\textwidth}
\centerline{
\begin{overpic}[trim =10 10 10 10, clip, width=1\textwidth]
{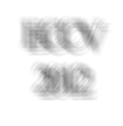}
\put(80, 30){ \includegraphics[width=0.05\textwidth]{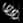}}
\put(2,3){\scriptsize\color{black}{\bf (a) Input}}
\end{overpic}}
\end{minipage}
\begin{minipage}[b]{.115\textwidth}
\centerline{
\begin{overpic}[trim =10 10 10 10, clip, width=1\textwidth]
{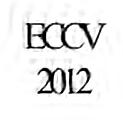}
\put(2,3){\scriptsize\color{black}{\bf (b) Levin \cite{levin2007image}}}
\end{overpic}}
\end{minipage}
\begin{minipage}[b]{.115\textwidth}
\centerline{
\begin{overpic}[trim =10 10 10 10, clip, width=1\textwidth]
{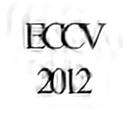}
\put(2,3){\scriptsize\color{black}{\bf (c) EPLL \cite{zoran2011epll}}}
\end{overpic}}
\end{minipage}
\begin{minipage}[b]{.115\textwidth}
\centerline{
\begin{overpic}[trim =10 10 10 10, clip,  width=1\textwidth]
{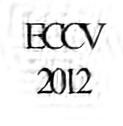}
\put(2,3){\scriptsize\color{black}{\bf (d) MLP \cite{schuler2013mlp}}}
\end{overpic}}
\end{minipage}
\begin{minipage}[b]{.115\textwidth}
\centerline{
\begin{overpic}[trim =10 10 10 10, clip,  width=1\textwidth]
{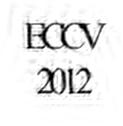}
\put(2,3){\scriptsize\color{black}{\bf (e) CSF \cite{schmidt2014csf}}}
\end{overpic}}
\end{minipage}
\begin{minipage}[b]{.115\textwidth}
\centerline{
\begin{overpic}[trim =18 18 18 18, clip, width=1\textwidth]
{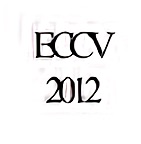}
\put(2,3){\scriptsize\color{black}{\bf \!(f) \!IRCNN\cite{zhang2017learning} }}
\end{overpic}}
\end{minipage}
\begin{minipage}[b]{.115\textwidth}
\centerline{
\begin{overpic}[trim =10 10 10 10, clip,  width=1\textwidth]
{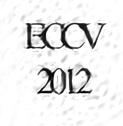}
\put(2,3){\scriptsize\color{black}{\bf (g) FDN \cite{kruse2017FFT}}}
\end{overpic}}
\end{minipage}
\begin{minipage}[b]{.115\textwidth}
\centerline{
\begin{overpic}[trim =0mm 0mm 0mm 0mm, clip, width=1\textwidth]
{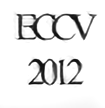}
\put(2,3){\scriptsize\color{black}{\bf (h) RGDN}}
\end{overpic}}
\end{minipage}
\vspace{-0.25cm}
\caption{Deconvolution results on a text image from \cite{pan2014text}.}
\label{fig:result_eccvtext}
\end{figure}

Fig. \ref{fig:result_eccvtext} shows a comparison on a text image where the blur kernel is from \cite{pan2014text}. The visual quality of the proposed method outperforms other methods, which implies that the proposed method can handle diverse images. The results of other methods suffer from heavy artifacts due to the imprecise blur kernel and unknown noise.

\par
\begin{figure*}[!t]
\vspace{-0.1cm}
  \centering
  \setlength{\tabcolsep}{1pt}
  \hspace{-0.1cm}
  \begin{tabularx}{0.96\textwidth}{cccccccccc}
    \multicolumn{2}{c}{\begin{overpic}[width=0.19\textwidth,trim=0
        8.7cm 0 0cm, clip]{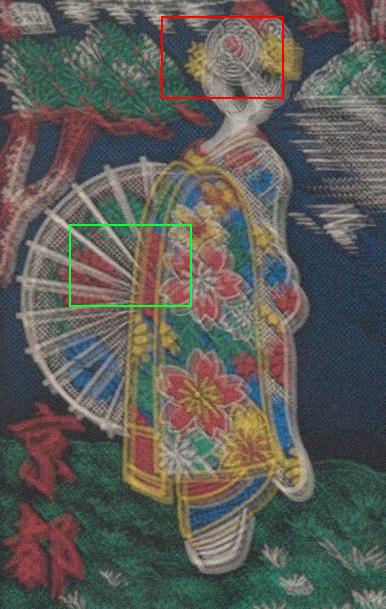}
      \put(78.75, 265){\includegraphics[width=0.04\textwidth]{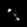}}
      \put(2,3){\footnotesize\color{white}{\bf (a) Input}}
    \end{overpic}}
    &
    \multicolumn{2}{c}{\begin{overpic}[width=0.19\textwidth,trim=0
      8.7cm 0 0cm,
      clip]{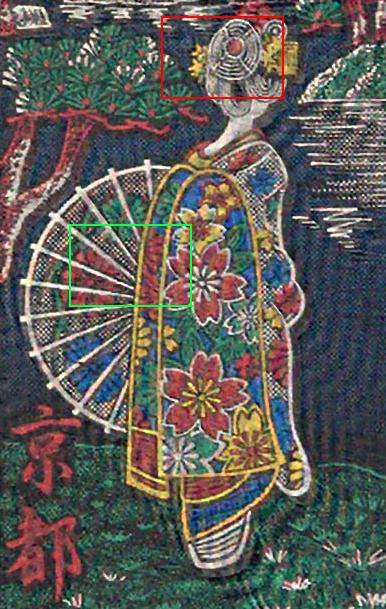}
      \put(2,3){\footnotesize\color{white}{\bf\!\! (b)\! Levin \etal\!\!\cite{levin2007image}}}
      \end{overpic}} &
      \multicolumn{2}{c}{\begin{overpic}[width=0.19\textwidth,trim=0
      8.7cm 0 0cm, clip]{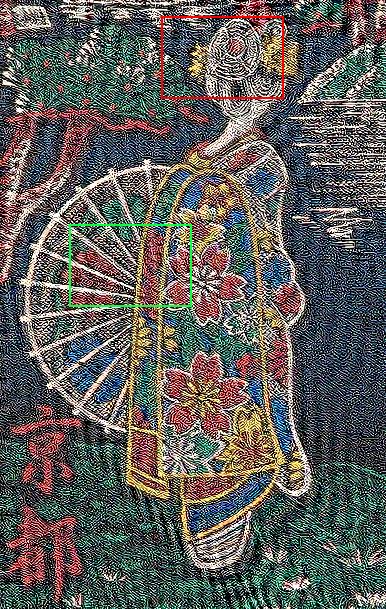}
      \put(2,3){\footnotesize\color{white}{\bf (c) IRCNN  \cite{zhang2017learning}}}
      \end{overpic}}
    &
      \multicolumn{2}{c}{\begin{overpic}[width=0.19\textwidth,trim=0
      8.7cm 0 0cm, clip]{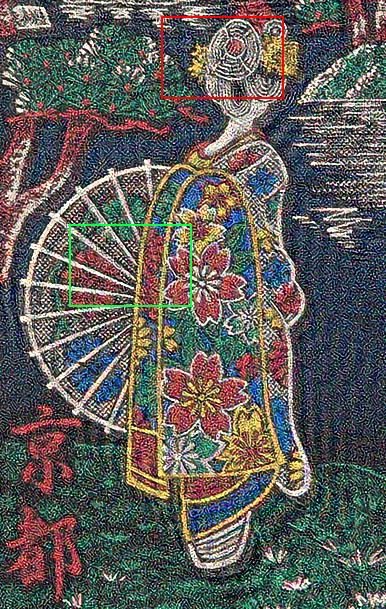}
      \put(2,3){\footnotesize\color{white}{\bf (d) FDN  \cite{kruse2017FFT}}}
      \end{overpic}}
    &
    \multicolumn{2}{c}{\begin{overpic}[width=0.19\textwidth,trim=0
      8.7cm 0 0cm, clip]{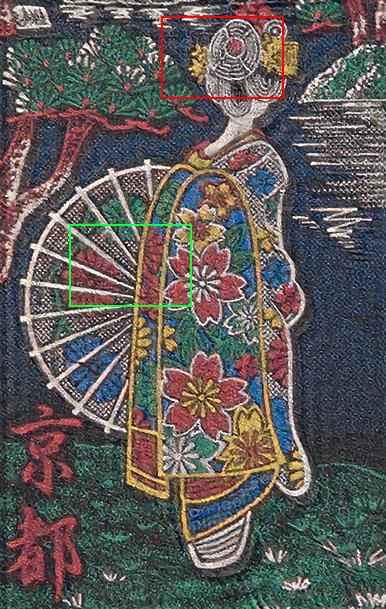}
      \put(2,3){\footnotesize\color{white}{\bf (e) RGDN (ours)}}
      \end{overpic}}\\
    \includegraphics[width=0.0925\textwidth]{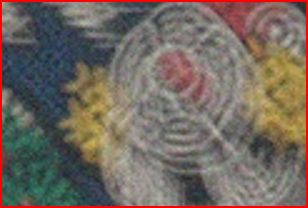}
    &
    \includegraphics[width=0.0925\textwidth]{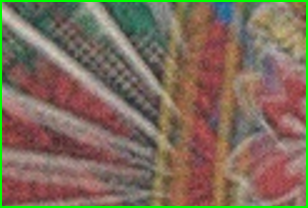}
    &
    \includegraphics[width=0.0925\textwidth]{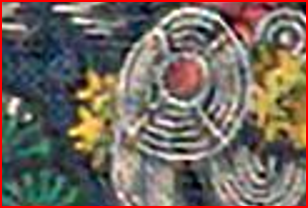}
    &
      \includegraphics[width=0.0925\textwidth]{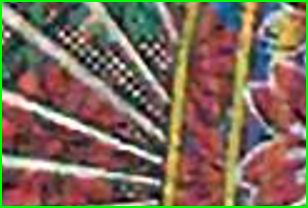}
    &
    \includegraphics[width=0.0925\textwidth]{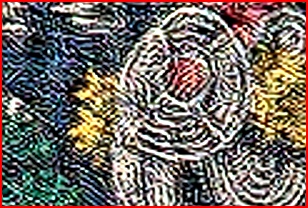}
    &
  \includegraphics[width=0.0925\textwidth]{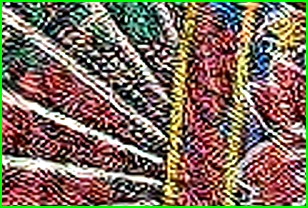}
    &
 \includegraphics[width=0.0925\textwidth]{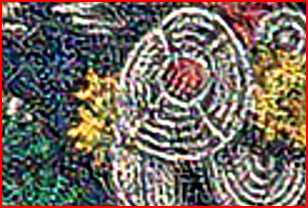}
    &
  \includegraphics[width=0.0925\textwidth]{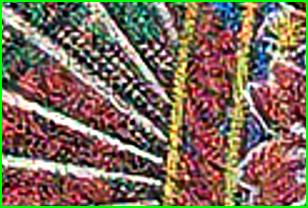}
    &
    \includegraphics[width=0.0925\textwidth]{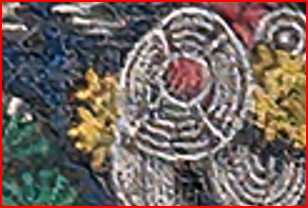}
    &
    \includegraphics[width=0.0925\textwidth]{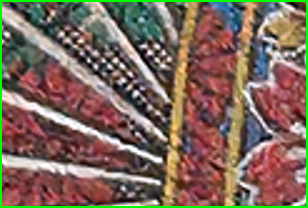}
  \end{tabularx}
  \vspace{-0.2cm}
   \caption{Deconvolution results on a real-world image with high level of noise. The images are best viewed by zooming in. }
  \label{fig:kyoto}
  \end{figure*}
\begin{figure*}[htp]
\centering
\begin{minipage}[b]{.24\textwidth}
\centerline{
\begin{overpic}[trim =10 10 10 10, clip, width=1\textwidth]
{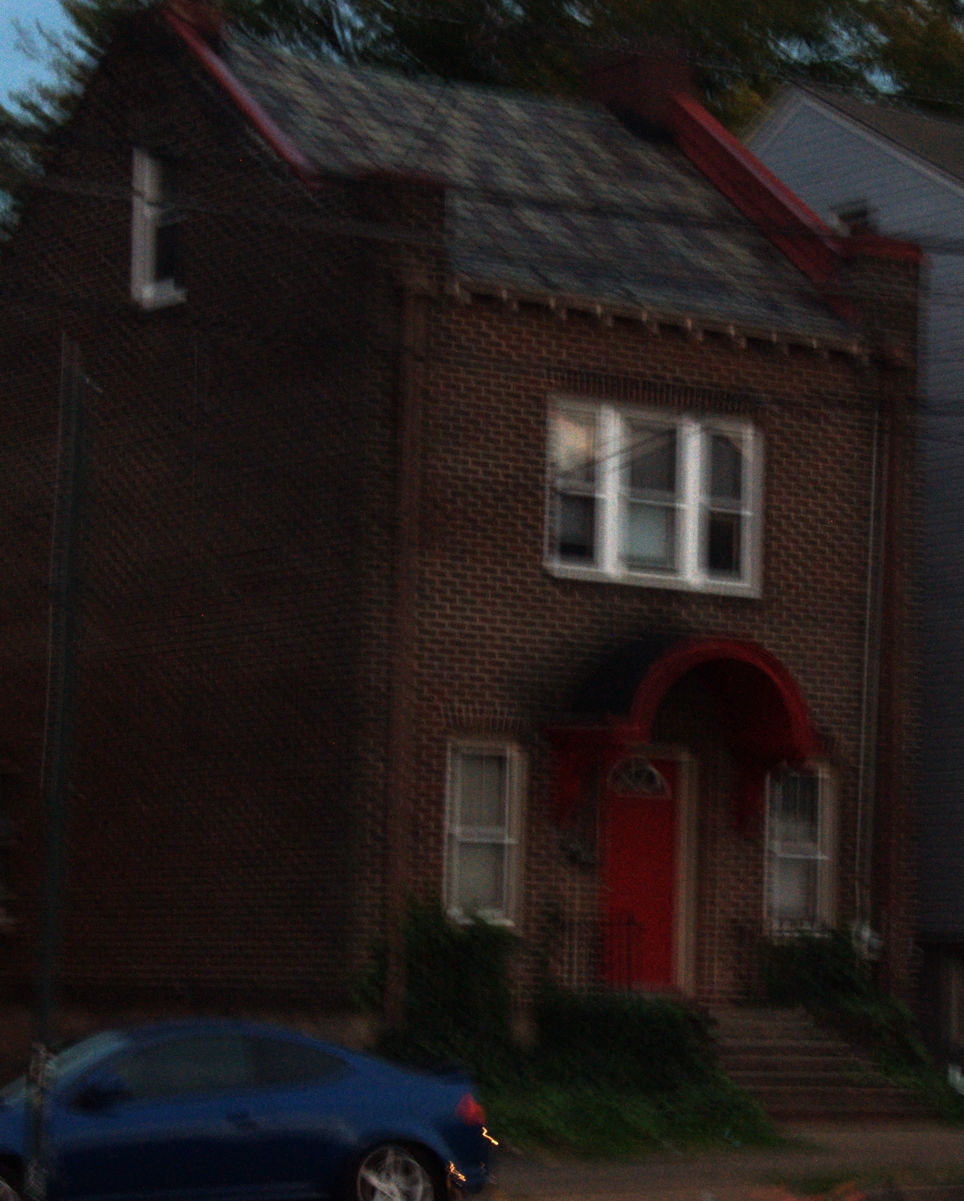}
\put(65, 87){ \includegraphics[width=0.1\textwidth]{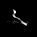}}
\put(2,3){\footnotesize\color{white}{\bf (a) Input}}
\end{overpic}}
\end{minipage}
\begin{minipage}[b]{.24\textwidth}
\centerline{
\begin{overpic}[trim =10 10 10 10, clip, width=1\textwidth]
{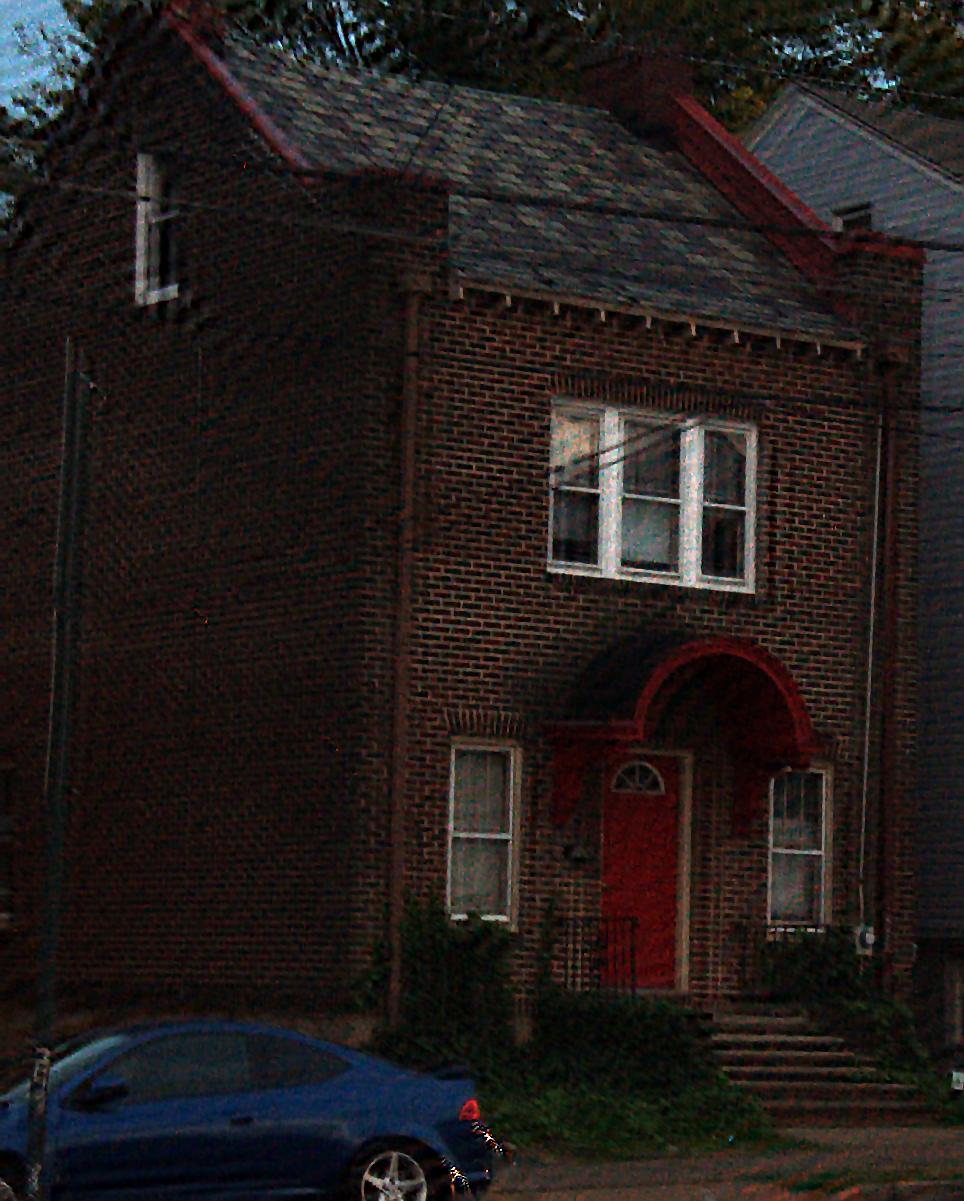}
\put(2,3){\footnotesize\color{white}{\bf (b) FD \cite{krishnan2009fast}} }
\end{overpic}}
\end{minipage}
\begin{minipage}[b]{.24\textwidth}
\centerline{
\begin{overpic}[trim =10 10 10 10, clip, width=1\textwidth]
{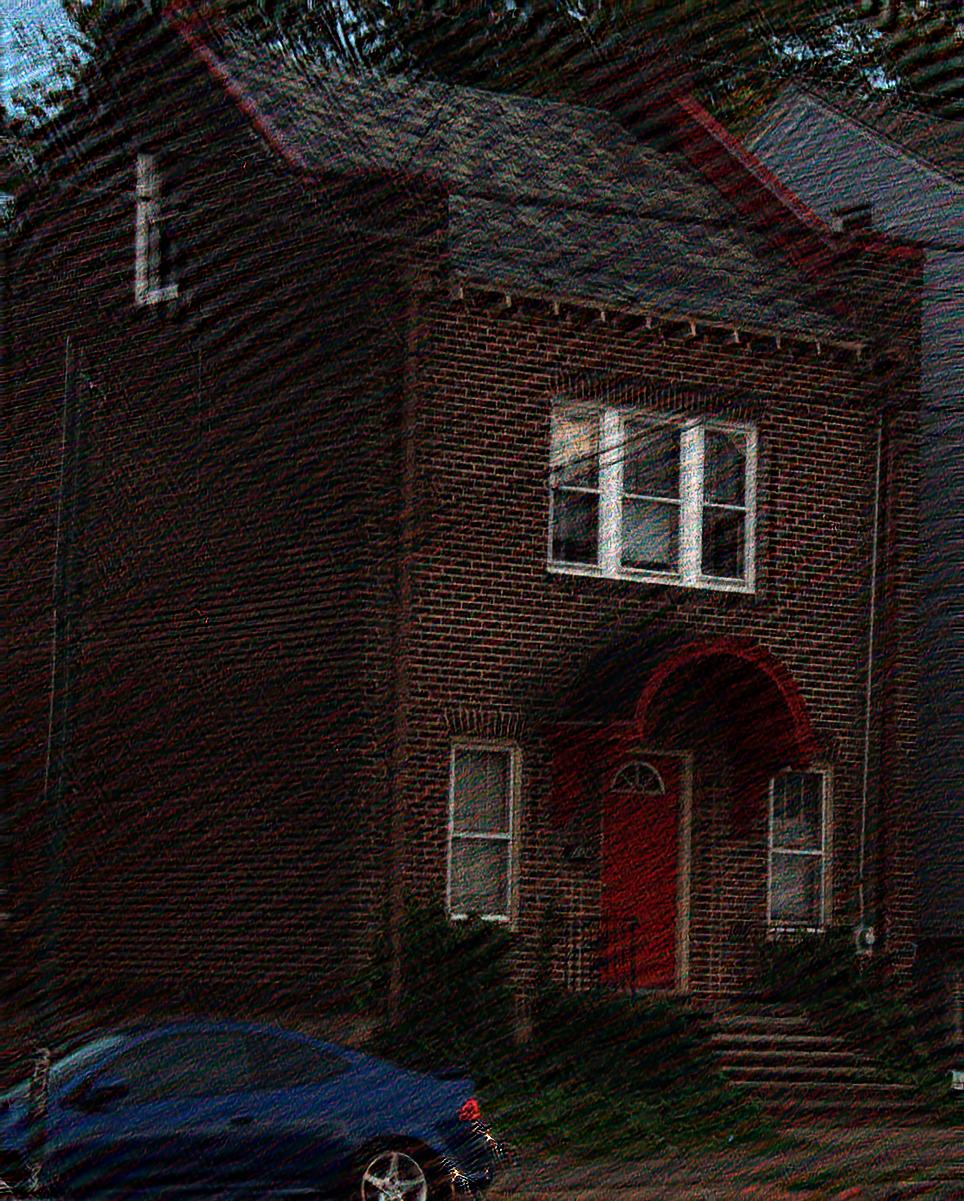}
\put(2,3){\footnotesize\color{white}{\bf (c) MLP \cite{schuler2013mlp}} }
\end{overpic}}
\end{minipage}
\begin{minipage}[b]{.24\textwidth}
\centerline{
\begin{overpic}[trim =10 10 10 10, clip, width=1\textwidth]
{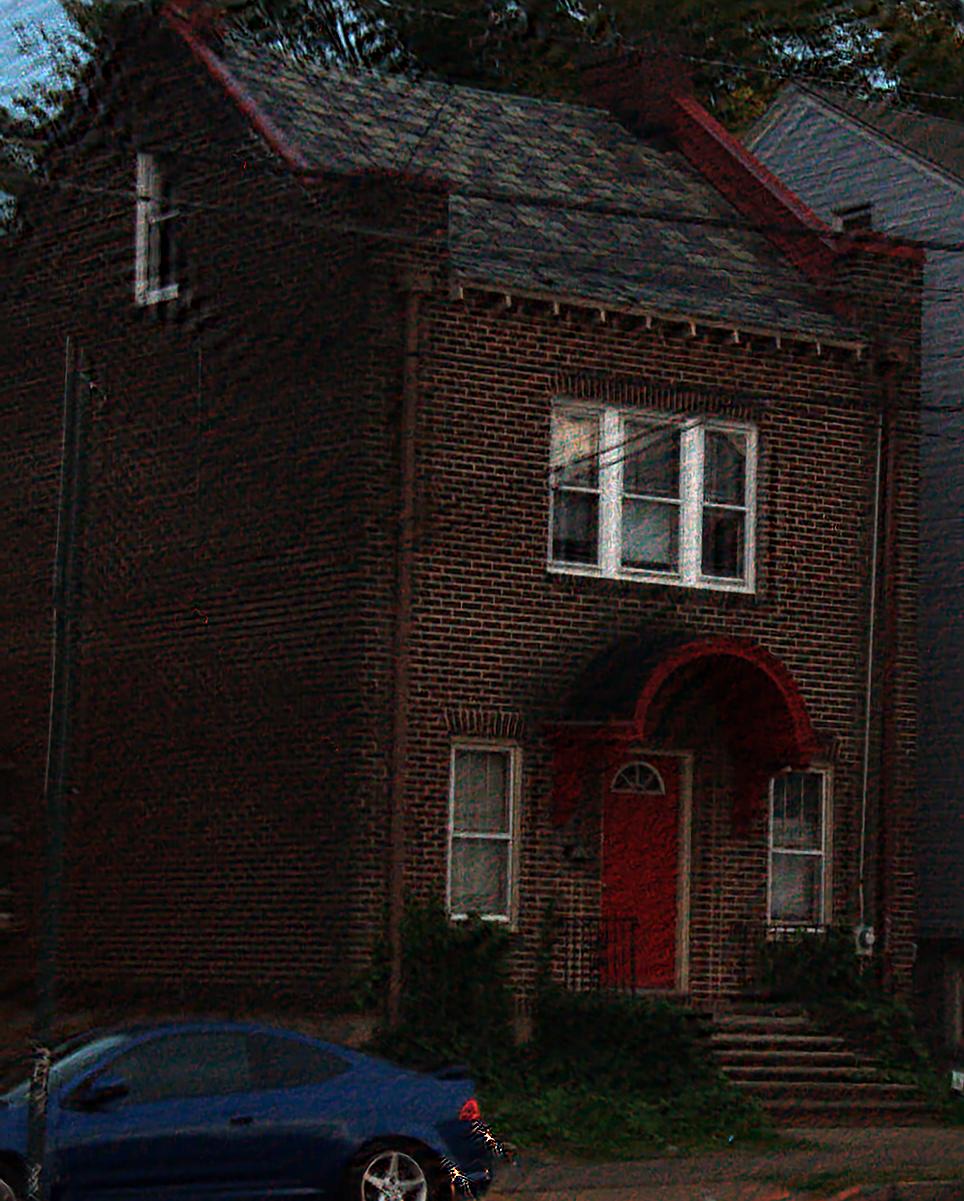}
\put(2,3){\footnotesize\color{white}{\bf (d) EPLL \cite{zoran2011epll} }}
\end{overpic}}
\end{minipage}\\
\vspace{0.1cm}
\begin{minipage}[b]{.24\textwidth}
\centerline{
\begin{overpic}[trim =10 10 10 10, clip, width=1\textwidth]
{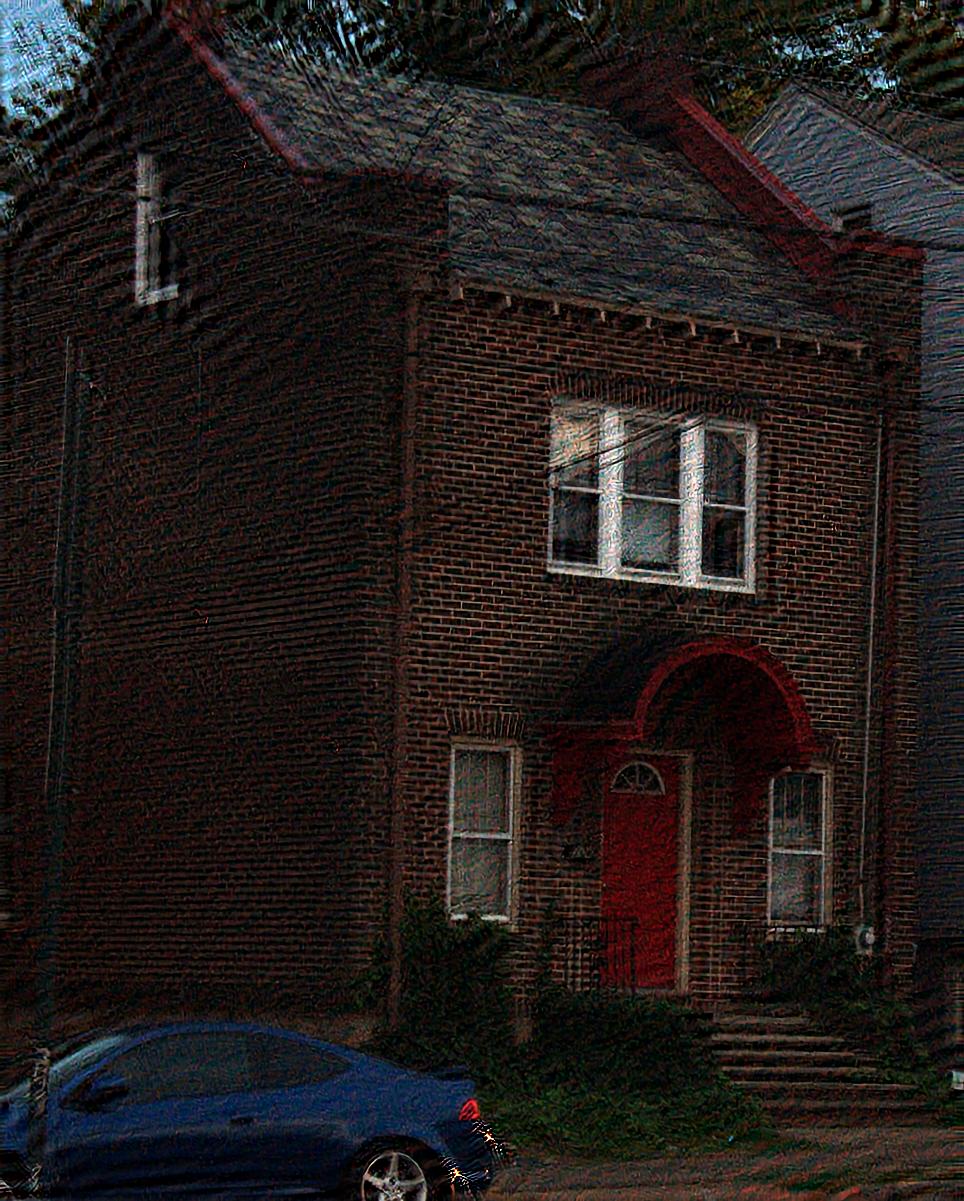}
\put(2,3){\footnotesize\color{white}{\bf (e) IRCNN \cite{zhang2017learning} }}
\end{overpic}}
\end{minipage}
\begin{minipage}[b]{.24\textwidth}
\centerline{
\begin{overpic}[trim =10 10 10 10, clip, width=1\textwidth]
{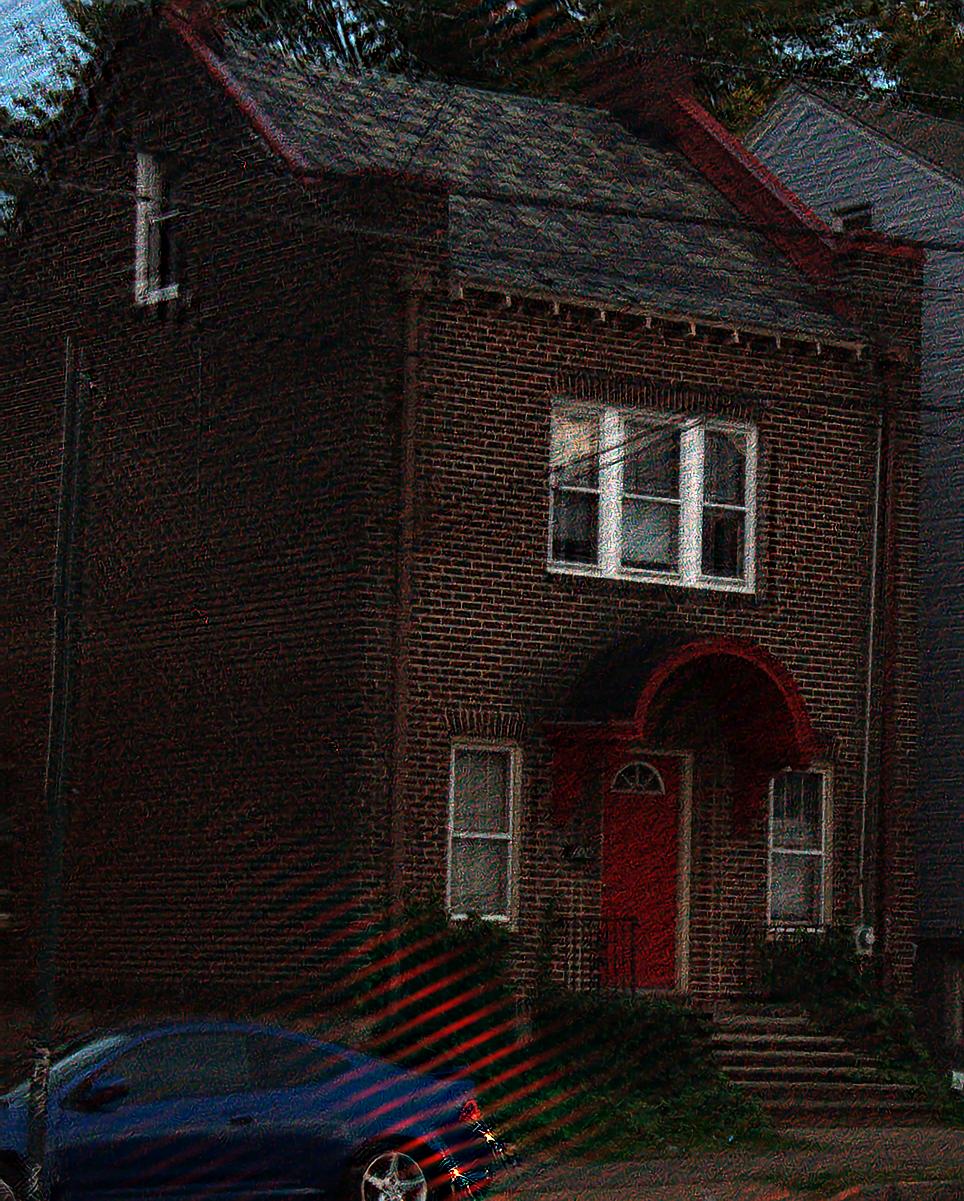}
\put(2,3){\footnotesize\color{white}{\bf (f) FDN \cite{kruse2017FFT}, $\sigma=0.59\%$ }}
\end{overpic}}
\end{minipage}
\begin{minipage}[b]{.24\textwidth}
\centerline{
\begin{overpic}[trim =10 10 10 10, clip, width=1\textwidth]
{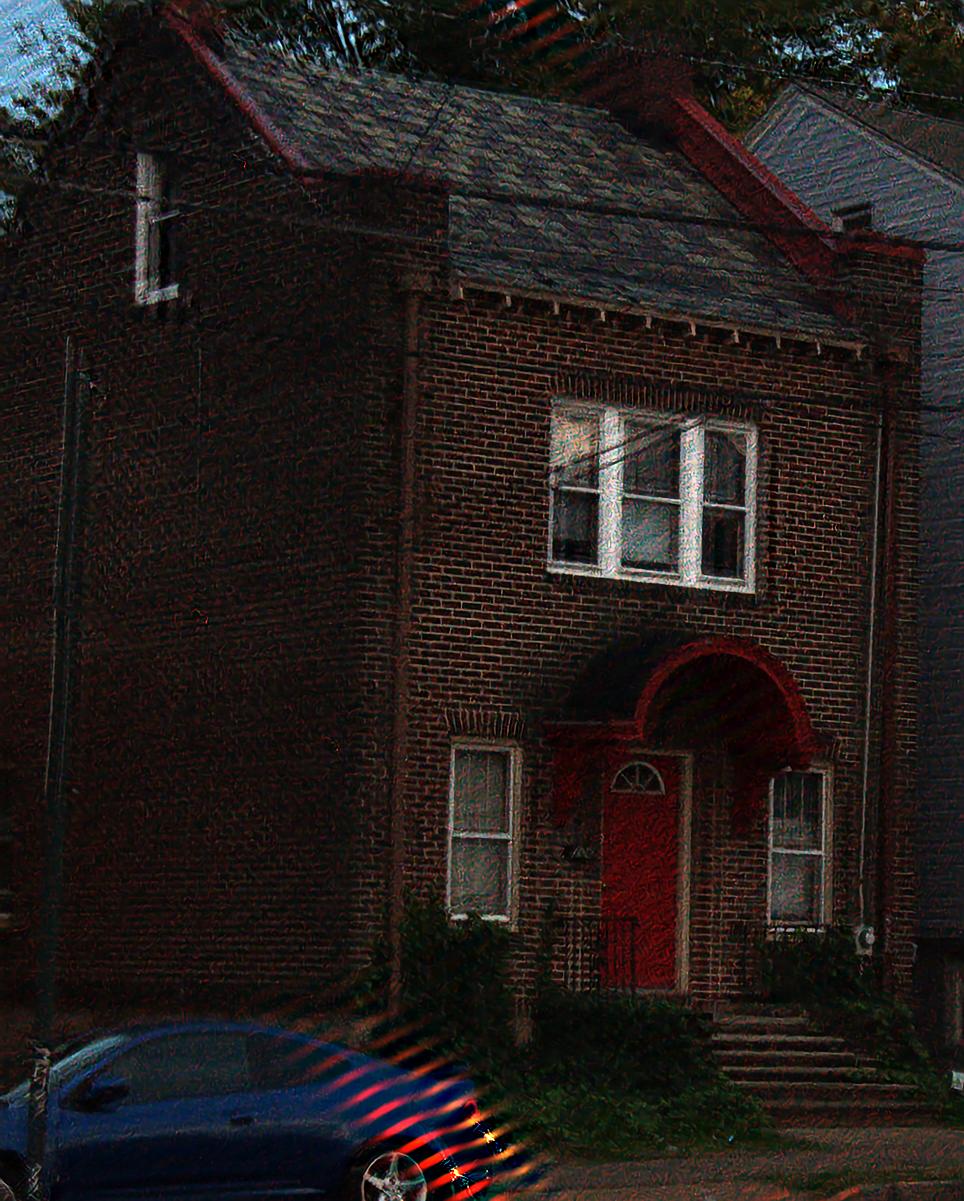}
\put(2,3){\footnotesize\color{white}{\bf (g) FDN \cite{kruse2017FFT}, $\sigma=1\%$}}
\end{overpic}}
\end{minipage}
\begin{minipage}[b]{.24\textwidth}
\centerline{
\begin{overpic}[trim =10 10 10 10, clip, width=1\textwidth]
{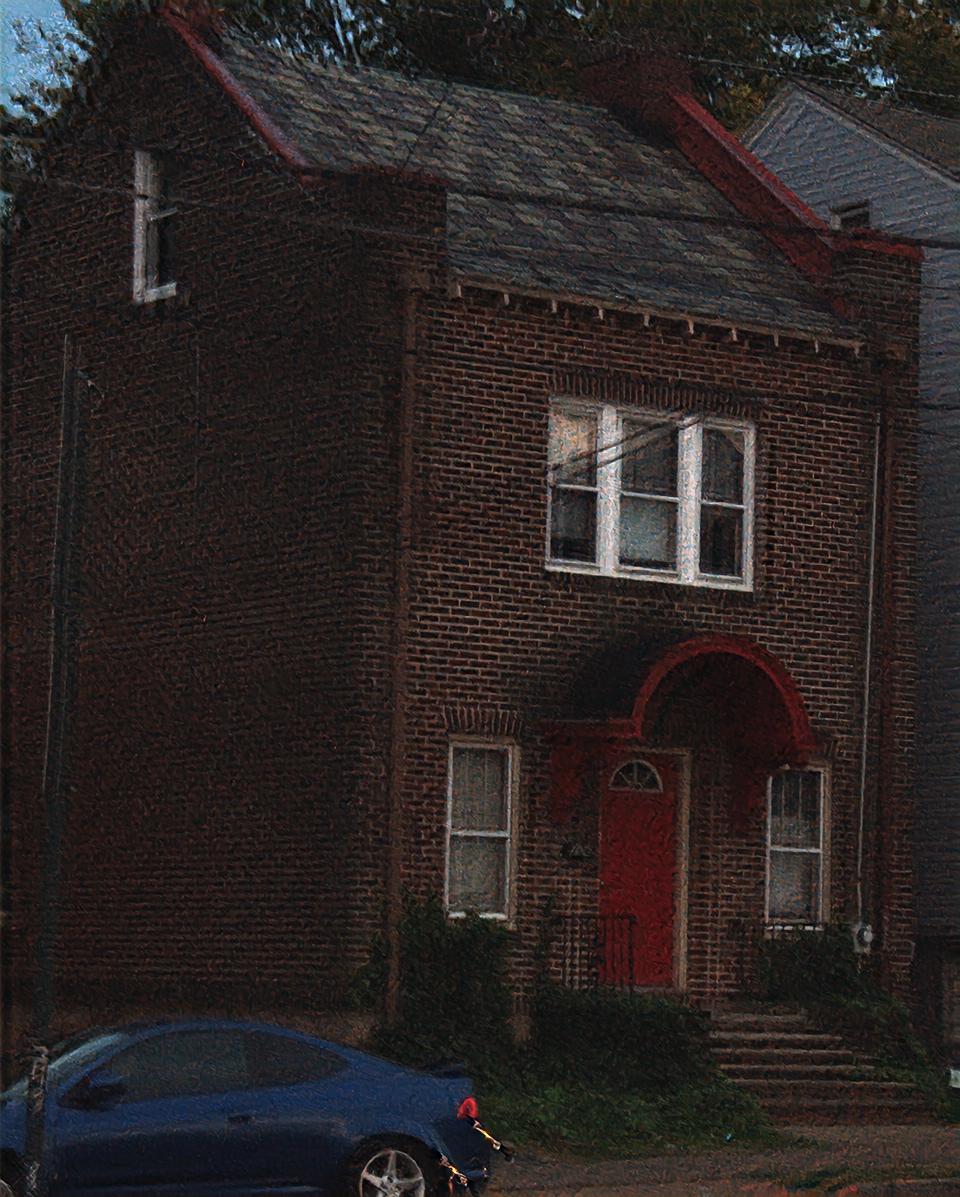}
\put(2,3){\footnotesize\color{white}{\bf (h) RGDN (ours)}}
\end{overpic}}
\end{minipage}
\caption{Deconvolution results on a real-world image with unknown noise and saturation. The images are best viewed by zooming in. Note that the proposed RGDN is free of hyper-parameter and the other methods require the noise level $\sigma$ as input. (f) and (g) show the results of FDN given different hyper-parameters. }
\label{fig:result_nb}
\end{figure*}

To assess the robustness of the propose method, we further test on a real-world blurry image with severe noise, in which the blur kernel is estimated using the method in \cite{gong2016blind}. As shown in Fig. \ref{fig:kyoto}, our restored image contains more details and suffers less ring artifacts than others. Fig. \ref{fig:kyoto} (c) shows that IRCNN \cite{zhang2017learning} does not competently handle the noise in the real-world image. Even though the result of FDN \cite{kruse2017FFT} may look sharp, it suffers from severe artifacts due the high noise level.
Fig. \ref{fig:result_nb} shows the experimental results of a real-world blurry image with unknown severe noise and saturations. The blur kernel is estimated by the method in \cite{gong2016blind}. The proposed RGDN recovers the image with sharp image and less artifacts. Due to the severe noise, the results of other methods suffer from the ringing artifacts or over-smoothed details. Fig. \ref{fig:result_nb} (f) and (g) show that the FDN \cite{kruse2017FFT} is sensitive to the saturation in the input image, resulting in strong ringing artifacts.
The proposed method is free of hyper-parameter. We tune the hyper-parameter (reflecting noise-level) for the compared methods and show results of FDN \cite{kruse2017FFT} given two different settings of the hyper-parameters. 

\par
{As shown in the visual comparisons on the real-world images, the results of the proposed method generally contain more details and fewer artifacts than other compared methods. We summarize the possible reasons in the following. 
Firstly, unlike previous related works \cite{kruse2017FFT,schmidt2014csf,zhang2017learning,zhang2016FCN} that merely model the image prior/regularization term, the proposed trainable optimizer also parameterizes and models the items corresponding to other components for handling the noise and boosting performances, \ie $\cH(\cdot)$ and $\cD(\cdot)$. 
Secondly, the proposed method shares the universal optimizer (and the model weights) across all steps. Thus, the final task is divided into a serious of small tasks in different stages, and the optimizer is trained to lift the image quality for each intermediate estimate dynamically, which is different to previous approaches statically training specific parameters for each step. The proposed model conforms to the iterative optimization method and can handle the dynamic optimization process better than other methods. 
Thirdly, parameter sharing, and the recursive supervision force the optimizer to handle various status in training. All the above characteristics make the proposed model general and robust to handle various samples, even the input blur kernel (\ie $\bk$ or $\bA$) may be inaccurate. 
}

\par
The experiments show that the proposed method is not only free of parameters and practical to use, but also robust to restore satisfactory results from the blurred images in different scenarios.

{
\subsection{Runtime and Memory Usage}
In this section, we report the runtime and memory usage of the proposed learned optimizer. While the time and memory usage of a specific method depends on the implementation and hardware device, we report the numbers of runtime and memory usage for the proposed method and some compared methods for an intuitive reference. 
}

\par
{
Our learned optimizer requires about 0.03 seconds and 1029MB GPU memory for one step on the small images (with $255\times 255\times 3$ pixels) of Levin \etal \cite{levin2009und}, and roughly 0.2 seconds and 5879MB memory for the samples (with $800\times 1024 \times 3$ pixels and $41\times 41$ kernels) from Sun \etal \cite{sun2013edge}. With the stopping conditions introduced above, the proposed optimizer (with 30 as the maximum iteration number) takes less than 1 second for each image from Levin \etal and about 5.8 seconds for each image from Sun \etal}

\par
{In experiments, we found the FFT based deep learning method FDN \cite{kruse2017FFT} uses 0.45 seconds for the small images from Levin \etal and 2.25 seconds for the larger images from Sun \etal\footnote{This number is for the 3-channel version of the data. The time reported in \cite{kruse2017FFT} is for single-channel gray image.}. 
Although the method in \cite{kruse2017FFT} is faster than the proposed method by using a fixed small number of iterations (not sharing parameters across steps), it is not as flexible and robust as the proposed model, as shown in above experiments. The other methods, such as Levin \cite{levin2009und} and EPLL \cite{zoran2011epll}, are much slower than the deep learning based methods, including the proposed method. Although it is not fair to compare the CPU based methods (\eg Levin \cite{levin2009und} and EPLL \cite{zoran2011epll}) with the GPU based implementations of the DNN methods, it implies the superiority of the DNN based methods on handling the realistic large images.}

\subsection{Extension to Image Denoising}
\label{sec:exp_denosing}
{Although the proposed method is mainly investigated for image deconvolution, we can extend the model straightforward to image denoising application. In this section, we show the potential of the proposed RGDN on image denosing. }
When the blur kernel $\bk$ in the imaging model \eqref{eq:blur_model} is a delta delta kernel $\bdelta$, \ie the corresponding convolution matrix $\bA$ in \eqref{eq:blur_model_matrix} is an identity matrix $\bI$, the observed image $\by$ is only degenerated by the noise term $\bn$ with $\by=\bx+\bn$. In this case, recovering $\bx$ from $\by$ becomes an image denoising problem.
By replacing $\bA$ in \eqref{eq:gd_comp_fun} with $\bI$, \ie changing the term $\bA^\T\bA\bx^t-\bA^\T\by$ as $\bx^t-\by$, we can easily adapt the proposed method for image denoising.

\par
We train an optimizer for denoising on the synthetic RGB images generated following the settings in Section \ref{sec:exp_setting}. 
Following the settings in \cite{zhang2017learning}, we evaluate the trained model using the on the (color) BSD68 dataset \cite{roth2005fields} with different noise levels.
Without loss of generality, we evaluate the model on the noisy images contaminated by Gaussian noise with different standard variance 
$\sigma=$25, 35 and 50\footnote{The noise levels 25, 35, 50 are corresponding to pixels in interval [0,255].}. 
{Table \ref{tab:denoising} shows the PSNR values of the denoising results of the classical method CBM3D \cite{egiazarian2015single}, learning based method TNRD \cite{chen2016trainable} and IRCNN \cite{zhang2017learning} and the proposed method. 
The proposed method is noise blind and free of hyper-parameters. 
However, the compared methods are noise non-blind and require the ground-truth noise level $\sigma$ as input hyper-parameter. 
As shown in Table \ref{tab:denoising}, the proposed method can achieve better results than CBM3D \cite{egiazarian2015single} and TNRD \cite{chen2016trainable}. 
Although the proposed method does not need to know the exact noise level, its performance is on par with the deep learning based denoiser in \cite{zhang2017learning}, which requires $\sigma$ as hyper-parameter. 
The results show the potential of the proposed method on the application beyond image deconvolution. 
}

\begin{table}[!t]
\centering
{
\caption{Evaluation of the Image Denoising Task on (color) BSD68 dataset. PSNR values are shown.}
\label{tab:denoising}
\begin{tabular}{llccc}
\toprule
\multicolumn{2}{c}{Noise level $\sigma$}    & 25 & 35 & 50 \\ \midrule
\multirow{3}{*}{Noise level non-blind} & CBM3D \cite{egiazarian2015single} &  30.71  & 28.89   &  27.38  \\
                                       & {TNRD \cite{chen2016trainable}} & {28.89} & {26.91} & {25.95}  \\
                                       & IRCNN \cite{zhang2017learning}    & 31.16      & 29.50      & 27.86 \\ \midrule
Noise level blind                  & RGDN (ours) & 30.94   &  29.29  &  27.66  \\ \bottomrule
\end{tabular}
}
\end{table}

\section{Conclusion}
We developed a Recurrent Gradient Descent Network (RGDN) which serves as an optimizer to de-convolute images. The components of the network are inspired by the key components of gradient descent method and designed accordingly.
The proposed RGDN implicitly learns a prior and tunes adaptive parameters through the CNN components of the gradient generator. The network has been trained on a diverse dataset thus is able to restore a wide range of blur images much better than previous approaches.
Our gradient descent unit is designed to handle Gaussian noise as specified in the $\ell_2$-loss (as shown in \eqref{eq:devon_min_prob}). One way to extend our network is to allow the gradient descent unit to model other type of noises or other losses.
{Considering the experiments in Section \ref{sec:exp_denosing} have shown the potential of the proposed method on the task beyond image deconvolution, we can also extend to proposed method to more image restoration applications. }

\ifCLASSOPTIONcaptionsoff
  \newpage
\fi

\bibliographystyle{IEEEtranbib}
\bibliography{learndeconv}

\begin{thebibliography}{10}
\providecommand{\url}[1]{#1}
\csname url@samestyle\endcsname
\providecommand{\newblock}{\relax}
\providecommand{\bibinfo}[2]{#2}
\providecommand{\BIBentrySTDinterwordspacing}{\spaceskip=0pt\relax}
\providecommand{\BIBentryALTinterwordstretchfactor}{4}
\providecommand{\BIBentryALTinterwordspacing}{\spaceskip=\fontdimen2\font plus
\BIBentryALTinterwordstretchfactor\fontdimen3\font minus
  \fontdimen4\font\relax}
\providecommand{\BIBforeignlanguage}[2]{{%
\expandafter\ifx\csname l@#1\endcsname\relax
\typeout{** WARNING: IEEEtran.bst: No hyphenation pattern has been}%
\typeout{** loaded for the language `#1'. Using the pattern for}%
\typeout{** the default language instead.}%
\else
\language=\csname l@#1\endcsname
\fi
#2}}
\providecommand{\BIBdecl}{\relax}
\BIBdecl

\bibitem{pan2014text}
J.~Pan, Z.~Hu, Z.~Su, and M.-H. Yang, ``Deblurring text images via
  l0-regularized intensity and gradient prior,'' in \emph{The IEEE Conference
  on Computer Vision and Pattern Recognition (CVPR)}, 2014, pp. 2901--2908.

\bibitem{xu2010two}
L.~Xu and J.~Jia, ``Two-phase kernel estimation for robust motion deblurring,''
  in \emph{European Conference on Computer Vision (ECCV)}, 2010, pp. 157--170.

\bibitem{gong2016motion}
D.~Gong, J.~Yang, L.~Liu, Y.~Zhang, I.~Reid, C.~Shen, A.~v.~d. Hengel, and
  Q.~Shi, ``From motion blur to motion flow: a deep learning solution for
  removing heterogeneous motion blur,'' \emph{The IEEE Conference on Computer
  Vision and Pattern Recognition (CVPR)}, 2017.

\bibitem{levin2007image}
A.~Levin, R.~Fergus, F.~Durand, and W.~T. Freeman, ``Image and depth from a
  conventional camera with a coded aperture,'' \emph{ACM transactions on
  graphics (TOG)}, vol.~26, no.~3, p.~70, 2007.

\bibitem{krishnan2009fast}
D.~Krishnan and R.~Fergus, ``Fast image deconvolution using hyper-laplacian
  priors,'' in \emph{Advances in Neural Information Processing Systems (NIPS)},
  2009, pp. 1033--1041.

\bibitem{wang2008new}
Y.~Wang, J.~Yang, W.~Yin, and Y.~Zhang, ``A new alternating minimization
  algorithm for total variation image reconstruction,'' \emph{SIAM Journal on
  Imaging Sciences}, vol.~1, no.~3, pp. 248--272, 2008.

\bibitem{zoran2011epll}
D.~Zoran and Y.~Weiss, ``From learning models of natural image patches to whole
  image restoration,'' in \emph{The IEEE International Conference on Computer
  Vision (ICCV)}, 2011, pp. 479--486.

\bibitem{schmidt2014csf}
U.~Schmidt and S.~Roth, ``Shrinkage fields for effective image restoration,''
  in \emph{The IEEE Conference on Computer Vision and Pattern Recognition
  (CVPR)}, 2014, pp. 2774--2781.

\bibitem{schuler2013mlp}
C.~J. Schuler, H.~Christopher~Burger, S.~Harmeling, and B.~Scholkopf, ``A
  machine learning approach for non-blind image deconvolution,'' in \emph{The
  IEEE Conference on Computer Vision and Pattern Recognition (CVPR)}, 2013, pp.
  1067--1074.

\bibitem{xu2014deep}
L.~Xu, J.~S. Ren, C.~Liu, and J.~Jia, ``Deep convolutional neural network for
  image deconvolution,'' in \emph{Advances in Neural Information Processing
  Systems (NIPS)}, 2014, pp. 1790--1798.

\bibitem{zhang2016FCN}
J.~Zhang, J.~Pan, W.-S. Lai, R.~Lau, and M.-H. Yang, ``Learning fully
  convolutional networks for iterative non-blind deconvolution,'' \emph{CVPR},
  2017.

\bibitem{chang2017one}
J.~R. Chang, C.-L. Li, B.~Poczos, B.~V. Kumar, and A.~C. Sankaranarayanan,
  ``One network to solve them all—solving linear inverse problems using deep
  projection models,'' 2017.

\bibitem{zhang2017learning}
K.~Zhang, W.~Zuo, S.~Gu, and L.~Zhang, ``Learning deep cnn denoiser prior for
  image restoration,'' in \emph{IEEE Conference on Computer Vision and Pattern
  Recognition (CVPR)}, 2017, pp. 3929--3938.

\bibitem{liu2018learning}
R.~Liu, X.~Fan, M.~Hou, Z.~Jiang, Z.~Luo, and L.~Zhang, ``Learning aggregated
  transmission propagation networks for haze removal and beyond,'' \emph{IEEE
  transactions on neural networks and learning systems}, no.~99, pp. 1--14,
  2018.

\bibitem{kruse2017FFT}
J.~Kruse, C.~Rother, and U.~Schmidt, ``Learning to push the limits of efficient
  fft-based image deconvolution,'' in \emph{IEEE International Conference on
  Computer Vision (ICCV)}, 2017, pp. 4586--4594.

\bibitem{schmidt2010generative}
U.~Schmidt, Q.~Gao, and S.~Roth, ``A generative perspective on mrfs in
  low-level vision,'' in \emph{The IEEE Conference on Computer Vision and
  Pattern Recognition (CVPR)}, 2013, pp. 1067--1074.

\bibitem{mptvgong}
D.~Gong, M.~Tan, Q.~Shi, A.~van~den Hengel, and Y.~Zhang, ``{MPTV}: Matching
  pursuit based total variation minimization for image deconvolution,''
  \emph{IEEE Transactions on Image Processing}, pp. 1--1, 2018.

\bibitem{goldstein2014fast}
T.~Goldstein, B.~O'Donoghue, S.~Setzer, and R.~Baraniuk, ``Fast alternating
  direction optimization methods,'' \emph{SIAM Journal on Imaging Sciences},
  vol.~7, no.~3, pp. 1588--1623, 2014.

\bibitem{sun2014good}
L.~Sun, S.~Cho, J.~Wang, and J.~Hays, ``Good image priors for non-blind
  deconvolution: Generic vs specific,'' in \emph{European Conference on
  Computer Vision (ECCV)}, 2014, pp. 231--246.

\bibitem{schmidt2016cascades}
U.~Schmidt, J.~Jancsary, S.~Nowozin, S.~Roth, and C.~Rother, ``Cascades of
  regression tree fields for image restoration,'' \emph{IEEE Transactions on
  Pattern Analysis and Machine Intelligence (TPAMI)}, vol.~38, no.~4, pp.
  677--689, 2016.

\bibitem{chen2015learning}
Y.~Chen, W.~Yu, and T.~Pock, ``On learning optimized reaction diffusion
  processes for effective image restoration,'' in \emph{The IEEE Conference on
  Computer Vision and Pattern Recognition (CVPR)}, 2015, pp. 5261--5269.

\bibitem{venkatakrishnan2013plug}
S.~V. Venkatakrishnan, C.~A. Bouman, and B.~Wohlberg, ``Plug-and-play priors
  for model based reconstruction,'' in \emph{Global Conference on Signal and
  Information Processing (GlobalSIP)}, 2013, pp. 945--948.

\bibitem{heide2016proximal}
F.~Heide, S.~Diamond, M.~Nie{\ss}ner, J.~Ragan-Kelley, W.~Heidrich, and
  G.~Wetzstein, ``Proximal: Efficient image optimization using proximal
  algorithms,'' \emph{ACM Transactions on Graphics (TOG)}, vol.~35, no.~4,
  p.~84, 2016.

\bibitem{geman1995nonlinear}
D.~Geman and C.~Yang, ``Nonlinear image recovery with half-quadratic
  regularization,'' \emph{IEEE Transactions on Image Processing}, vol.~4,
  no.~7, pp. 932--946, 1995.

\bibitem{jin2017noiseblind}
M.~Jin, S.~Roth, and P.~Favaro, ``Noise-blind image deblurring,'' in \emph{The
  IEEE Conference on Computer Vision and Pattern Recognition (CVPR)}, 2017.

\bibitem{li2016learning}
K.~Li and J.~Malik, ``Learning to optimize,'' \emph{arXiv preprint
  arXiv:1606.01885}, 2016.

\bibitem{andrychowicz2016learning}
M.~Andrychowicz, M.~Denil, S.~Gomez, M.~W. Hoffman, D.~Pfau, T.~Schaul, and
  N.~de~Freitas, ``Learning to learn by gradient descent by gradient descent,''
  in \emph{Advances in Neural Information Processing Systems (NIPS)}, 2016, pp.
  3981--3989.

\bibitem{ravi2016optimization}
S.~Ravi and H.~Larochelle, ``Optimization as a model for few-shot learning,''
  in \emph{International Conference on Learning Representations (ICLR)}, 2017.

\bibitem{zheng2015conditional}
S.~Zheng, S.~Jayasumana, B.~Romera-Paredes, V.~Vineet, Z.~Su, D.~Du, C.~Huang,
  and P.~H. Torr, ``Conditional random fields as recurrent neural networks,''
  in \emph{International conference on computer vision (ICCV)}, 2015, pp.
  1529--1537.

\bibitem{kobler2017variational}
E.~Kobler, T.~Klatzer, K.~Hammernik, and T.~Pock, ``Variational networks:
  connecting variational methods and deep learning,'' in \emph{German
  conference on pattern recognition}, 2017, pp. 281--293.

\bibitem{klatzer2016learning}
T.~Klatzer, K.~Hammernik, P.~Knobelreiter, and T.~Pock, ``Learning joint
  demosaicing and denoising based on sequential energy minimization,'' in
  \emph{International Conference on Computational Photography (ICCP)}, 2016,
  pp. 1--11.

\bibitem{gu2018integrating}
S.~Gu, R.~Timofte, and L.~Van~Gool, ``Integrating local and non-local denoiser
  priors for image restoration,'' in \emph{International Conference on Pattern
  Recognition (ICPR)}, 2018, pp. 2923--2928.

\bibitem{gu2017learning}
S.~Gu, W.~Zuo, S.~Guo, Y.~Chen, C.~Chen, and L.~Zhang, ``Learning dynamic
  guidance for depth image enhancement,'' in \emph{IEEE Conference on Computer
  Vision and Pattern Recognition (CVPR)}, 2017, pp. 3769--3778.

\bibitem{wieschollek2017learning}
P.~Wieschollek, M.~Hirsch, B.~Sch{\"o}lkopf, and H.~Lensch, ``Learning blind
  motion deblurring,'' \emph{The IEEE International Conference on Computer
  Vision (ICCV)}, 2017.

\bibitem{kim2017online}
T.~H. Kim, K.~M. Lee, B.~Sch{\"o}lkopf, and M.~Hirsch, ``Online video
  deblurring via dynamic temporal blending network,'' \emph{arXiv preprint
  arXiv:1704.03285}, 2017.

\bibitem{kim2016deeply}
J.~Kim, J.~Kwon~Lee, and K.~Mu~Lee, ``Deeply-recursive convolutional network
  for image super-resolution,'' in \emph{The IEEE Conference on Computer Vision
  and Pattern Recognition (CVPR)}, 2016, pp. 1637--1645.

\bibitem{liu2016recfilter}
S.~Liu, J.~Pan, and M.-H. Yang, ``Learning recursive filters for low-level
  vision via a hybrid neural network,'' in \emph{European Conference on
  Computer Vision (ECCV)}, 2016, pp. 560--576.

\bibitem{parikh2014proximal}
N.~Parikh, S.~Boyd \emph{et~al.}, ``Proximal algorithms,'' \emph{Foundations
  and Trends{\textregistered} in Optimization}, vol.~1, no.~3, pp. 127--239,
  2014.

\bibitem{gong2017mpgl}
D.~Gong, M.~Tan, Y.~Zhang, A.~van~den Hengel, and Q.~Shi, ``{MPGL}: An
  efficient matching pursuit method for generalized lasso.'' in \emph{AAAI},
  2017, pp. 1934--1940.

\bibitem{wright1999numerical}
S.~Wright and J.~Nocedal, ``Numerical optimization,'' \emph{Springer Science},
  vol.~35, no. 67-68, p.~7, 1999.

\bibitem{ioffe2015bn}
S.~Ioffe and C.~Szegedy, ``Batch normalization: Accelerating deep network
  training by reducing internal covariate shift,'' in \emph{International
  Conference on Machine Learning (ICML)}, 2015, pp. 448--456.

\bibitem{fan2017generic}
Q.~Fan, J.~Yang, G.~Hua, B.~Chen, and D.~Wipf, ``A generic deep architecture
  for single image reflection removal and image smoothing,'' in
  \emph{Proceedings of the IEEE International Conference on Computer Vision
  (ICCV)}, 2017.

\bibitem{kinga2015adam}
D.~Kinga and J.~B. Adam, ``A method for stochastic optimization,'' in
  \emph{International Conference on Learning Representations (ICLR)}, 2015.

\bibitem{sun2013edge}
L.~Sun, S.~Cho, J.~Wang, and J.~Hays, ``Edge-based blur kernel estimation using
  patch priors,'' in \emph{IEEE International Conference on Computational
  Photography (ICCP)}, 2013, pp. 1--8.

\bibitem{pytorch}
\url{https://github.com/pytorch/pytorch}.

\bibitem{Everingham15}
M.~Everingham, S.~M.~A. Eslami, L.~Van~Gool, C.~K.~I. Williams, J.~Winn, and
  A.~Zisserman, ``The pascal visual object classes challenge: A
  retrospective,'' \emph{International Journal of Computer Vision (IJCV)}, vol.
  111, no.~1, pp. 98--136, 2015.

\bibitem{chakrabarti2016neural}
A.~Chakrabarti, ``A neural approach to blind motion deblurring,'' in
  \emph{European Conference on Computer Vision (ECCV)}, 2016, pp. 221--235.

\bibitem{levin2009und}
A.~Levin, Y.~Weiss, F.~Durand, and W.~T. Freeman, ``Understanding and
  evaluating blind deconvolution algorithms,'' in \emph{The IEEE Conference on
  Computer Vision and Pattern Recognition (CVPR)}, 2009, pp. 1964--1971.

\bibitem{chen2016trainable}
Y.~Chen and T.~Pock, ``Trainable nonlinear reaction diffusion: A flexible
  framework for fast and effective image restoration,'' \emph{IEEE transactions
  on pattern analysis and machine intelligence (TPAMI)}, vol.~39, no.~6, pp.
  1256--1272, 2016.

\bibitem{sun2012super}
L.~Sun and J.~Hays, ``Super-resolution from internet-scale scene matching,'' in
  \emph{IEEE International Conference on Computational Photography (ICCP)},
  2012, pp. 1--12.

\bibitem{bsd500}
P.~Arbelaez, M.~Maire, C.~Fowlkes, and J.~Malik, ``Contour detection and
  hierarchical image segmentation,'' \emph{IEEE Transactions on Pattern
  Analysis and Machine Intelligence (TPAMI)}, vol.~33, no.~5, pp. 898--916,
  2011.

\bibitem{chambolle2010introduction}
A.~Chambolle, V.~Caselles, D.~Cremers, M.~Novaga, and T.~Pock, ``An
  introduction to total variation for image analysis,'' \emph{Theoretical
  foundations and numerical methods for sparse recovery}, vol.~9, no. 263-340,
  p. 227, 2010.

\bibitem{wang2004ssim}
Z.~Wang, A.~C. Bovik, H.~R. Sheikh, and E.~P. Simoncelli, ``Image quality
  assessment: from error visibility to structural similarity,'' \emph{IEEE
  Transactions on Image Processing}, vol.~13, no.~4, pp. 600--612, 2004.

\bibitem{gong2016blind}
D.~Gong, M.~Tan, Y.~Zhang, A.~Van~den Hengel, and Q.~Shi, ``Blind image
  deconvolution by automatic gradient activation,'' in \emph{The IEEE
  Conference on Computer Vision and Pattern Recognition (CVPR)}, 2016, pp.
  1827--1836.

\bibitem{gong2017self}
D.~Gong, M.~Tan, Y.~Zhang, A.~van~den Hengel, and Q.~Shi, ``Self-paced kernel
  estimation for robust blind image deblurring,'' in \emph{Proceedings of the
  IEEE Conference on Computer Vision and Pattern Recognition}, 2017, pp.
  1661--1670.

\bibitem{roth2005fields}
S.~Roth and M.~J. Black, ``Fields of experts: A framework for learning image
  priors,'' in \emph{IEEE Conference on Computer Vision and Pattern
  Recognition}, vol.~2, 2005, pp. 860--867.

\bibitem{egiazarian2015single}
K.~Egiazarian and V.~Katkovnik, ``Single image super-resolution via bm3d sparse
  coding,'' in \emph{Signal Processing Conference (EUSIPCO), 2015 23rd
  European}, 2015, pp. 2849--2853.

\end{thebibliography}

\end{document}